\newif\ifanonymous

\documentclass[runningheads]{llncs}
\usepackage{graphicx}
\usepackage{comment}
\usepackage{amsmath,amssymb} 
\usepackage{color}

\ifanonymous
\usepackage{ruler}
\usepackage[width=122mm,left=12mm,paperwidth=146mm,height=193mm,top=12mm,paperheight=217mm]{geometry}
\fi

\usepackage{url}            
\usepackage{booktabs}       
\usepackage{nicefrac}       
\usepackage{microtype}      

\usepackage{float}
\usepackage[inline]{enumitem}

\usepackage{multirow}
\usepackage{cite}
\usepackage{algorithm}
\usepackage[noend]{algpseudocode}

\usepackage[pagebackref=true,breaklinks=true,letterpaper=true,colorlinks,bookmarks=false]{hyperref}
\hypersetup{
	colorlinks=true,
	linkcolor=magenta,
	filecolor=green,      
	urlcolor=magenta,
	citecolor=cyan,
}

\def\etal{\emph{et al}.\ }
\newcommand{\region}{\mathbb{R}}
\newcommand{\bool}{\mathbb{B}}
\DeclareMathOperator*{\expectation}{\mathbb{E}}

\DeclareMathOperator*{\Bernoulli}{\text{Bern}}

\def\ECCVSubNumber{2109} 	

\begin{document}

\pagestyle{headings}
\mainmatter

\title{Out-of-the-box Channel Pruned Networks} 

\ifanonymous
\titlerunning{Channel Pruning \ECCVSubNumber} 
\authorrunning{ECCV-20 submission ID \ECCVSubNumber} 
\author{Anonymous ECCV submission}
\institute{Paper ID \ECCVSubNumber}
\else
\titlerunning{Channel Pruning}
\author{Ragav Venkatesan\inst{1, 2}\and
	Gurmurthy Swaminathan\inst{1} \and
	Xiong Zhou\inst{1} \and
	Runfei Luo\inst{1}}
\authorrunning{R. Venkatesan et al.}
\institute{Amazon Web Services AI \and
	Amazon Alexa AI \\
	\email{\{ragavven, gurumurs, xiongzho, annaluo\}@amazon.com}}
\fi
\maketitle

\begin{abstract}
	In the last decade convolutional neural networks have become gargantuan. 
	Pre-trained models, when used as initializers are able to fine-tune ever larger networks on small datasets.
	Consequently, not all the convolutional features that these fine-tuned models detect are requisite for the end-task. 
	Several works of channel pruning have been proposed to prune away compute and memory from models that were trained already. 
	Typically, these involve policies that decide which and how many channels to remove from each layer leading to channel-wise and/or layer-wise pruning profiles, respectively. 
	In this paper, we conduct several baseline experiments and establish that profiles from random channel-wise pruning policies are as good as metric-based ones.
	We also establish that there may exist profiles from some layer-wise pruning policies that are measurably better than common baselines.
	We then demonstrate that the top layer-wise pruning profiles found using an exhaustive random search from one datatset are also among the top profiles for other datasets.
	This implies that we could identify out-of-the-box layer-wise pruning profiles using benchmark datasets and use these directly for new datasets.
	Furthermore, we develop a Reinforcement Learning (RL) policy-based search algorithm with a direct objective of finding transferable layer-wise pruning profiles using many models for the same architecture.
	We use a novel reward formulation that drives this RL search towards an expected compression while maximizing accuracy.
	Our results show that our transferred RL-based profiles are as good or better than best profiles found on the original dataset via exhaustive search.
	We then demonstrate that if we found the profiles using a mid-sized dataset such as Cifar10/100, we are able to transfer them to even a large dataset such as Imagenet.
\end{abstract}

\section{Introduction}
\label{sec:intro}

Channel Pruning is the process of removing entire convolutional channels while trying to not raise the error of a CNN substantially.
Consider a convolutional neural network $N$ with $l$ layers.
Consider that $N$ is trained with dataset $D_b$ producing an artifact $N_b$.
Suppose that the weights of these layers $w_t  \in \region^{k_h \times k_w \times c_{t-1} \times c_t}, \ t \in \{1, \dots , l\}$ and $c_0$  is the number of channels of the input images.
Consider the input of the $t^{\text{th}}$ layer, $L_t \in \region^{h \times w \times c_{t-1}} $ and its output  $L_{t+1} \in \region^{h \times w \times c_t}$ such that $L_{t} = L_{t-1} * w_t$. 
To prune channels in $N_b$, we need a \textit{pruning profile} that is a set of Boolean vectors, one per-layer each, with a dimensionality the same as the number of channels in the layer,
\begin{equation}
p  = \{  \boldsymbol{\alpha_i} \}, \boldsymbol{\alpha_i} \in \bool^{c_i} , \forall i \in \{ 1, \dots, l\}.
\label{eqn:genereal_channel_profile}
\end{equation}
Once we have such a profile, we can perform channel pruning as follows,
\begin{equation}
L_t = L_{t-1} * ( \boldsymbol{\alpha_{t-1}} \circ_3 w_t \circ_4 \boldsymbol{\alpha_t} ), \forall t \in \{1, \dots l\},
\end{equation}
where, $\circ_i$ is a broadcasted Hadamard product on the $i^\text{th}$ dimension and $\boldsymbol{\alpha_0} = \boldsymbol{1}$.
Simply put, $\boldsymbol{\alpha_{t-1}}$ prunes the filter (or input) dimension of $w_t$ and $\boldsymbol{\alpha_t}$ prunes the channel (or output) dimension of $w_t$.
Accordingly, the $\boldsymbol{\alpha}$ that prunes the channels of a layer, also prunes the filters of the subsequent layer.

Channel pruning revolves around policies that can produce profiles similar to $p$.
Consider now the pruning of layer $i$ using a policy $\pi$.
The policy $\pi$ is simply a function that takes some layer features $\Phi(L_i, w_i)$ as input (such as mean activations or $\ell_1$ of each channels etc.,) and outputs a Boolean vector, with each dimension denoting whether to prune or to retain the corresponding channel.

\begin{equation}
\pi(\Phi(L_i, w_i)) =  \boldsymbol{\alpha_i},  \boldsymbol{\alpha_i} \in \bool^{c_i}
\end{equation}
The pruning profile for pruning such the network $N_b$ extracted from $\pi$ is,
\begin{equation}
p  = \{ \pi( \Phi(L_i, w_i))\} = \{  \boldsymbol{\alpha_i} \}, \forall i \in \{ 1, \dots, l\}
\label{eqn:all_profile}
\end{equation}

This policy searches for how many channels to remove and which particular channels to remove in a network simultaneously.
Good profiles in the form of equation \ref{eqn:all_profile} are impractical to find by brute-force or dynamic-programming as, such a search space of approximately $\bool^{\max\{c_i\} \times l}$ is monumentally large.
However, we can apply several relaxations, that could be used to produce smaller search spaces.
There are generally two common types of relaxations that we consider leading to layer-wise pruning policies and channel-wise pruning policies.

Consider the channel-wise pruning policy $\pi_c$:
\begin{equation}
\pi_c(\Phi(L_i, w_i))=\boldsymbol{\alpha_i}, \text{ such that } \boldsymbol{\alpha_i} \in \bool^{c_i} \text{ subject to } \sum_1^{c_i} \boldsymbol{\alpha_i} = k_i, k_i \in [0, 1]
\label{eqn:channel_wise}
\end{equation}
where, $k_i$ is a fixed heuristic that represents how many channels to remove every layer.
Note that the search space of  $\pi_c$ is the same as $\pi$, even though $\pi_c$ is operating with a narrower set of possible solutions.
Channel-wise policies typically operate under a fixed assumption of how much to prune each layer.
Consider now the following example for a layer-wise pruning policy $\pi_l$:
\begin{equation}
\pi_l(\Phi(L_i, w_i)) = \boldsymbol{\alpha_i},  \boldsymbol{\alpha_i} =  [\Bernoulli(\beta_i)], \beta_i \in [0,1], i \in \{1, \dots c_i \}.
\label{eqn:layer_wise}
\end{equation}
The layer-wise pruning profile for pruning $N_b$ using $\pi_l$ is,
\begin{equation}
p_l = \{ \pi_l(\Phi(L_i, w_i))\}  = \{\beta_i\}, \ \in \{1, \dots, l\}.
\end{equation}
This is a much simpler search space in $\region^l$.
In this case, $\beta_i$ can be considered as the probability of pruning any channel in layer $L_i$. 
Consequently, the policy is determining how many channels to remove each layer and randomly deciding which channels to remove in each layer.

In this work, we derive some insights on different types of policies and argue that layer-wise policies are preferable to channel-wise policies.
We also introduce the concept of \textit{out-of-the-box pruned networks}, which are networks that come with pre-identified pruning profiles.
Particularly, we demonstrate that top layer-wise profiles found using a base network $N_b$ (trained on $D_b$) are also among the top profiles for $N_t$ (trained on $D_t$), out-of-the-box.
We then develop a Reinforcement Learning (RL) search-based system to identify transferable profiles that perform on unseen target network-datasets $N_t$ similar to or better than top profiles discovered directly using $N_t$. 

In this work, we use the terms \textit{base network} and \textit{pretrained initialization} to refer to the original network and its weights after the initial training and prior to being pruned.
For our experimentation, we choose to use three networks: 
\begin{enumerate}[wide, labelwidth=!, labelindent=0pt, nosep,before=\leavevmode]
	\item \textbf{C-NET:} A special case of the AlexNet philosophy~\cite{krizhevsky2010convolutional}. 
	C-NET or CylinderNet is a network with the same number of channels in every layer.
	\item \textbf{Resnet-20-16, Resnet-20-64:} Resnet-20-16 is a Resnet of 20 layers with 16, 32 and 128 channels on each of its blocks and Resnet-20-64 is a Resnet of 20 layers with 64, 128 and 512 channels on each of its blocks, respectively~\cite{he2016deep}\footnote{For more details on these networks, and pruning the shortcut and non-convolutional layers in Resnet, refer to the supplementary.}. 
\end{enumerate}
We use \textbf{Cifar10, Cifar100, TinyImagenet and Imagenet} datasets for various experiments~\cite{krizhevsky2009learning, tiny, deng2009imagenet}.

In this work we use compression factor (CF) as a way to measure degree of compression.
We define compression factor as $\text{CF} = \frac{\vert w \vert}{\vert w^p \vert}$, where $\text{CF}$ is typically expressed in x-\emph{factor} as in, 2x, 3x and so on, $ \vert w^p \vert$ is the number of parameters left in the pruned network and $\vert w \vert$ is the number of parameters in the base network.  
Using such a CF helps us cover a meaningful span of compression albeit on a non-linear scale. Anytime when there is randomness involved in the profiles, we run them 5 times and have presented mean and standard deviations accordingly.

\subsection{Contributions}
The contributions of this paper are as follows:
\begin{enumerate}[wide, labelwidth=!, labelindent=0pt, nosep,before=\leavevmode]
	\item \textbf{Layer-wise pruning profiles matter the most:}
	Firtsly, we study both random and trained network weights as options for initialization during fine-tuning and conclude that given a well-optimized fine-tuning process, the plasticity of the neural networks assists in recovering performance even from random initialization.
	We also find that in some cases, random initialization is better than the trained weights as initialization for fine-tuning.
	While Liu \etal verified this for unstructured pruning, we independently verify this for channel pruning~\cite{liu2018rethinking}.
	For channel-wise pruning policies, we notice that popular baselines from literature such as selecting weights with least $\ell_1$ or Taylor features show noticeable benefit, but random choices are as good in most conditions.
	Indeed, we argue that in the case of computational constraints, random choice as channel-wise policy to prune is sufficient. 
	While Mittal \etal and Qin \etal argue this for unstructured pruning, we independently verify this for channel pruning. 
	Additionally, we also show that under heavy compression max-metric methods do have a noticeable benefit~\cite{mittal2018recovering, qin2018demystifying}.
	When it comes to layer-wise pruning policies, we demonstrate that the popular baseline of equally-distributing compression across all layers as a heuristic is wanting and that the optimal pruning profile is architecture-dependent. 
	Equally-distributed profiles are a popular baseline because they retain the original shape of the architecture. 
	This makes them a good profile for lighter compression, but under heavier compression, we demonstrate that there exist better architecture-specific optimal-pruning profiles.
	We also establish that at all CFs, the performance of different profiles varies significantly implying that layer-wise profiles do matter.
	
	\item \textbf{Good pruning profiles remain good even when transferred:} We demonstrate that layer-wise pruning profiles are transferable across datasets. 
	Consider that we identify a top channel pruning profile $p^b$ using a base network $N_b$.
	We could then apply the same pruning profile $p^b$ on another unseen network $N_t$ and, expect it to be as good as top profiles $p^t$ identified directly using $N_t$.
	So long as $N_b$ and $N_t$ are the same architecture $N$, we discover that profiles are transferable, out-of-the-box.
	To the  best of our knowledge, there exists no prior work that discusses and analyzes the transferability of channel pruning profiles.
	
	\item \textbf{RL-driven search yields better transferable profiles:} We propose a novel RL-based approach to search for better transferable channel pruning profiles. 
	In particular, we develop a multi-environment setup, where we train one pruning policy on a set of networks $N_B = \{N_i\}, i \in \{1,\dots n\}$ and produce pruning profiles $p^B = \{p_i\}, i \in \{1,\dots n\}$. 
	We find that the profiles $p^B$ are among the top profiles $p^t$ identified using $N_{t} \notin N_B$.
	To the best of our knowledge, this is the first work that performs generalizable and transferable channel pruning with (either a random search or) a RL-based approach, even though  there are some prior works that train RL-based policies for pruning a particular network~\cite{zoph2016neural, ashok2018nn, he2018amc, cai2017reinforcement}.
\end{enumerate}

The rest of the paper is organized as follows: section \ref{sec:taxonomy} establishes our initial premises, section \ref{sec:layerwise} demonstrates the value of layer-level pruning policies and the transferability of top randomly searched pruning profiles, section \ref{sec:case} introduces the novel RL-based, transferable pruning profile search algorithm and demonstrates its value and, section \ref{sec:conclusions} provides concluding remarks.

\section{Establishing Premises}
\label{sec:taxonomy}

A pruning policy could either be deterministic, learned online (during pruning for the particular network) or a general policy that was learned offline, say a policy for all Resnets~\cite{he2016deep}.
Often, parts of policies are learned online or offline while others are deterministic.
While there are policies that iteratively prune more and more at every stage, we do not consider the iterative process to be that of learning the policy.
Therefore works such as those by Hu \etal can be considered deterministic~\cite{hu2016network, venkatesan2016diving}.
Works such as those by  He \etal, Huang \etal, Wen \etal, Alvarez \etal, and Ye \etal are considered online even though the policy is embedded into the network via regularization~\cite{he2017iccv, Huang_2018_ECCV,  NIPS2016_6504, NIPS2016_6372, ye2018rethinking}.
Works such as those by Lin \etal on train-time pruning are also categorized similarly, as they are simply regularized-training where the regularization process is to reshape the network architecture~\cite{lin2017nips}.
We also do not consider unstructured pruning applied to convolutional networks and focus only on channel pruning~\cite{NIPS2015_5784}.

\subsection{On the strategies for re-initialization during fine-tuning.}

Any network once pruned, has to be fine-tuned in order to recover the original accuracy of the model.
Typically, the base network's weights are used as initialization in this phase.
Recently, several works have suggested the use of rewound weights from the original network as initializations~\cite{frankle2018lottery, zhou2019deconstructing, morcos2019one}.
In a demurral of the lottery ticket hypothesis, some works have suggested that initializing the weights from scratch works just as well~\cite{liu2018rethinking, zhou2019deconstructing}.
While there is no prior work that learns to generate weights for reinitialization, some methods do learn to adapt original pre-trained weights along with learning the pruning policies themselves~\cite{molchanov2017variational, louizos2017learning}.
These can be considered as fine-tuning as well.
The axiomatic way to initialize in literature appears to be from final trained weights of the base network as-is.

Consider the heuristic of pruning every channel in the network with an equal probability $k$ using the policy,
\begin{equation}
\pi_r(c_i) = \{{\{{\Bernoulli(k)\}}_{\times c_i}\}}_{\times l} \text{ and } k \in [0,1].
\label{eqn:random_channel}
\end{equation}
We refer to $\pi_r$ as ``equally-distributed random policy" and the profile $\pi_r$ produces an ``equally-distributed random pruning profile" .
This means that for every layer, we remove the same percentage of channels but which channels we remove is chosen at random.
The experiment setup is as follows:
Firstly, we train base models for all networks on all datasets.
We then sweep over various CF using the profile above.
After pruning each network, we fine-tune the pruned network with two strategies:
\begin{enumerate*}
	\item reinitialize $w$ from scratch, which we refer to as ``random initialization" and
	\item initialize $w$ for those channels that remain, the weights from the base network $N_b$, which we refer to as `pretrained intialization'.
\end{enumerate*}
We train several pruned models at various CFs.

\begin{figure}[htb!]
	\includegraphics[width=0.325\linewidth]{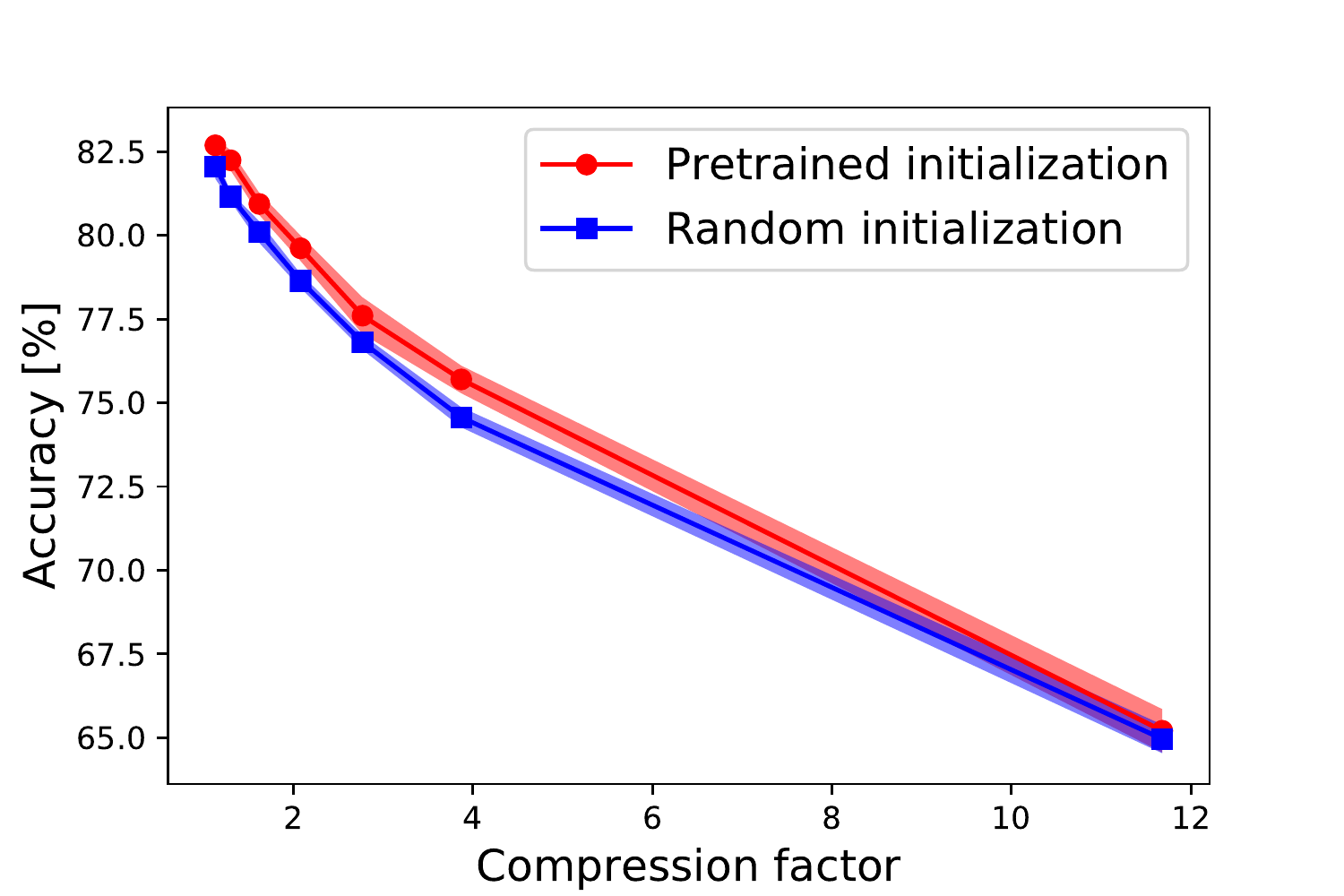}
	\includegraphics[width=0.325\linewidth]{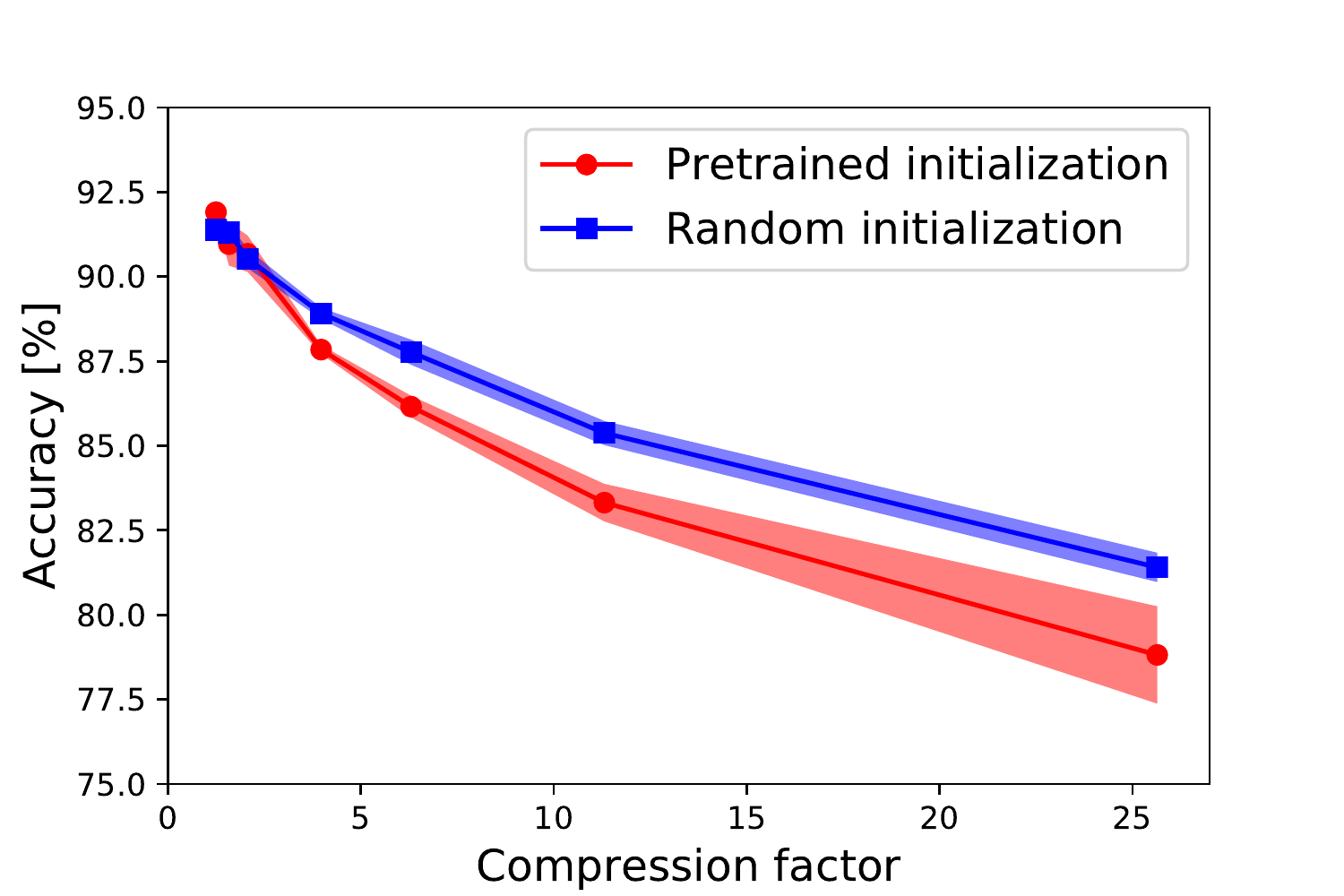}
	\includegraphics[width=0.325\linewidth]{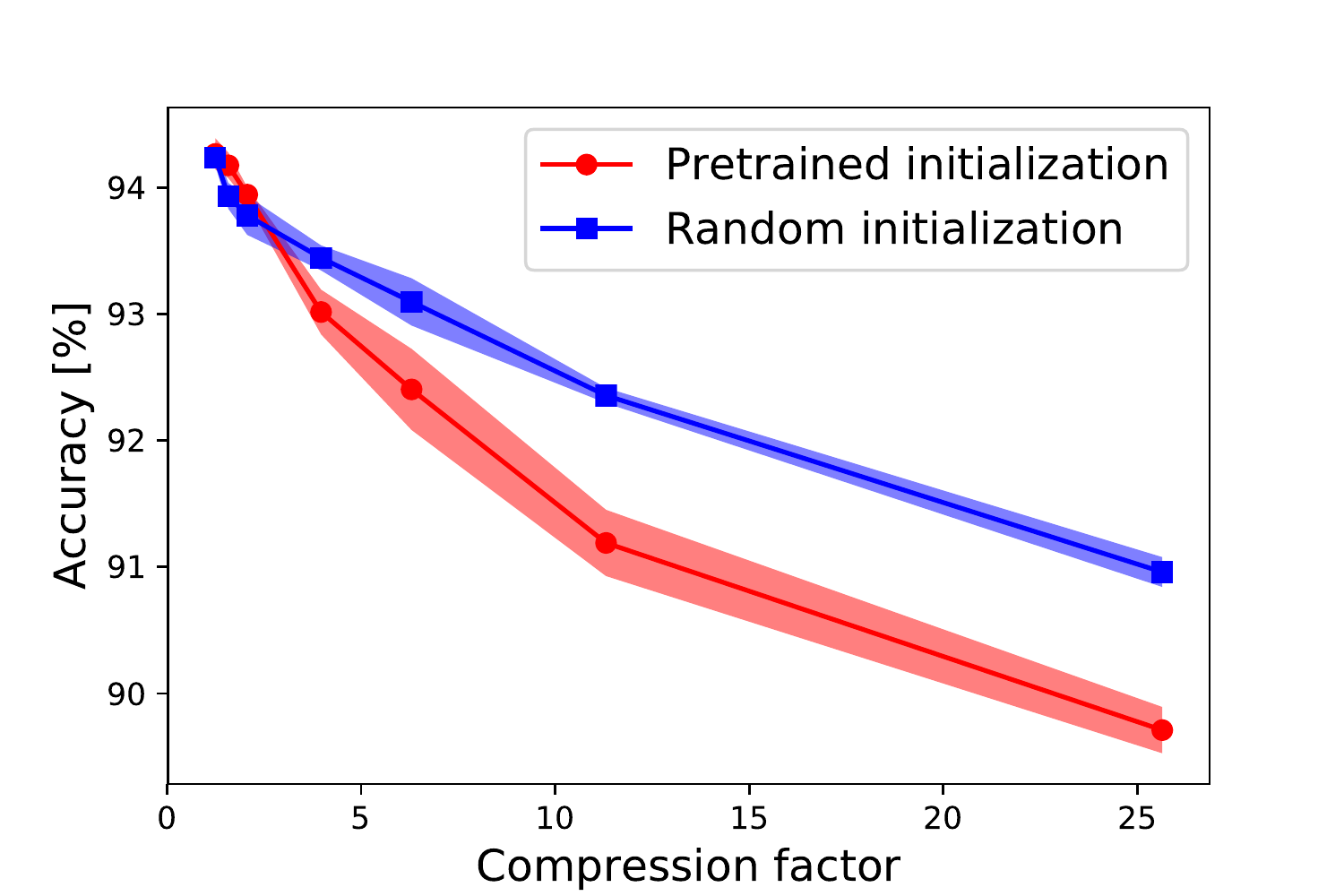}
	\includegraphics[width=0.325\linewidth]{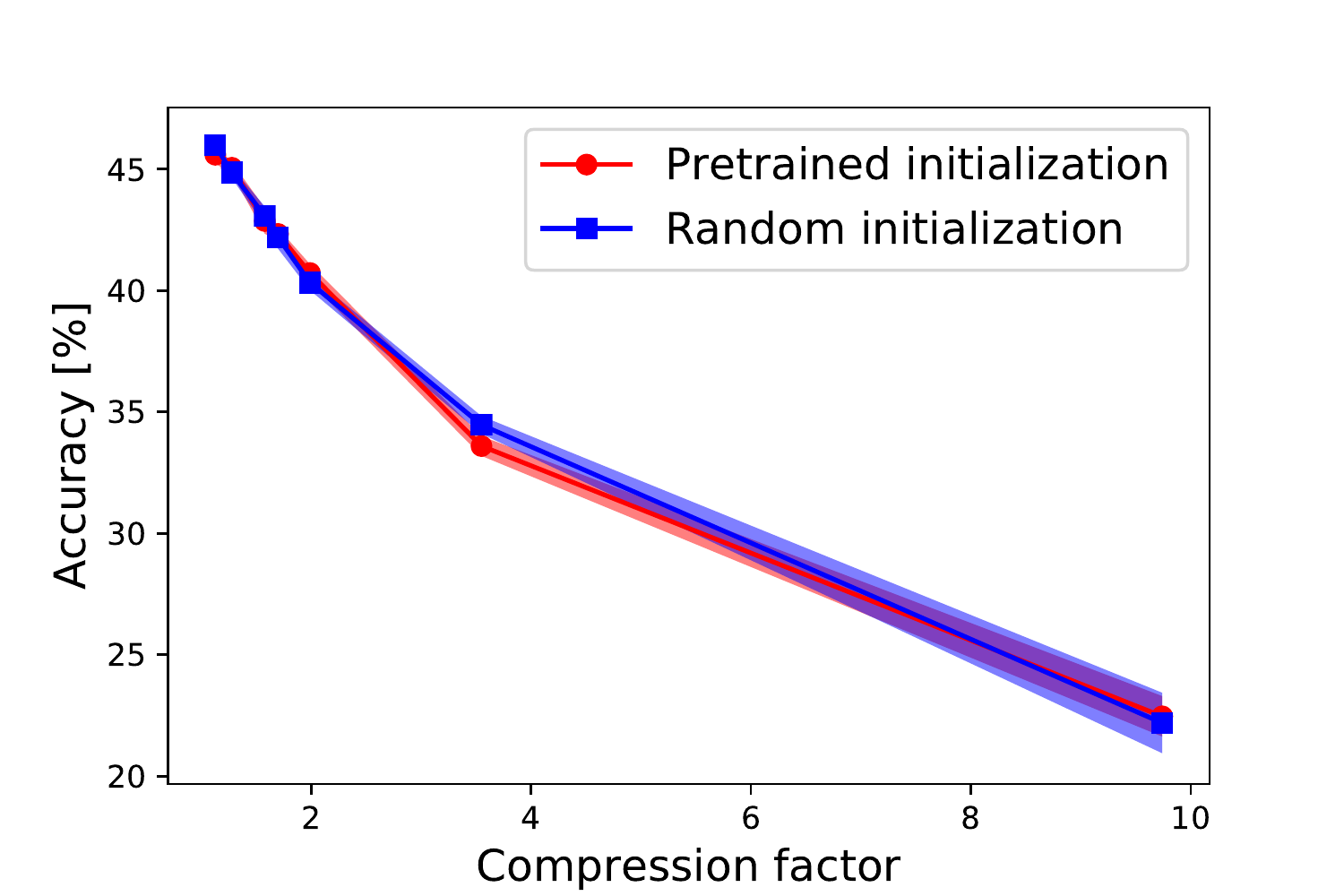}
	\includegraphics[width=0.325\linewidth]{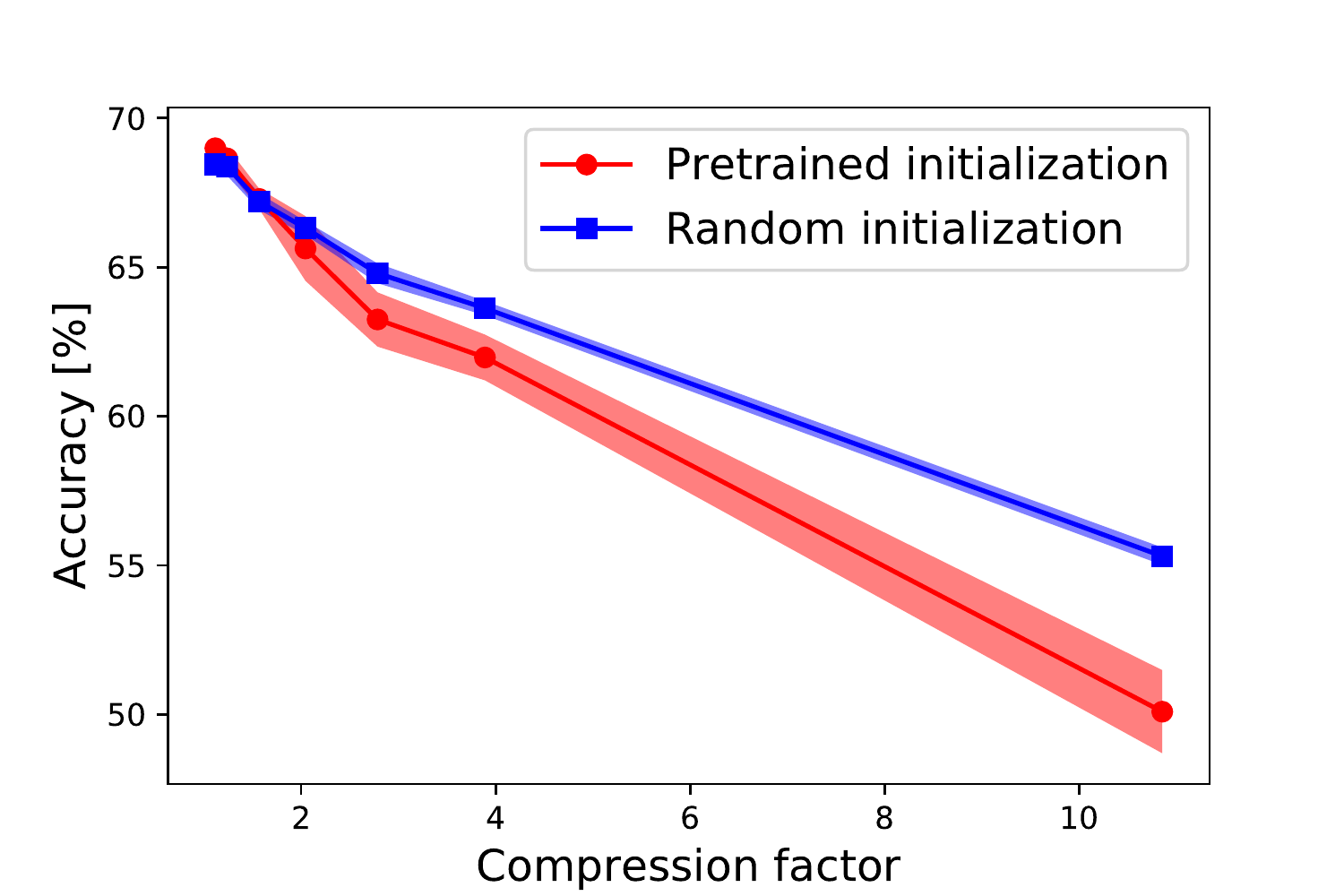}
	\includegraphics[width=0.325\linewidth]{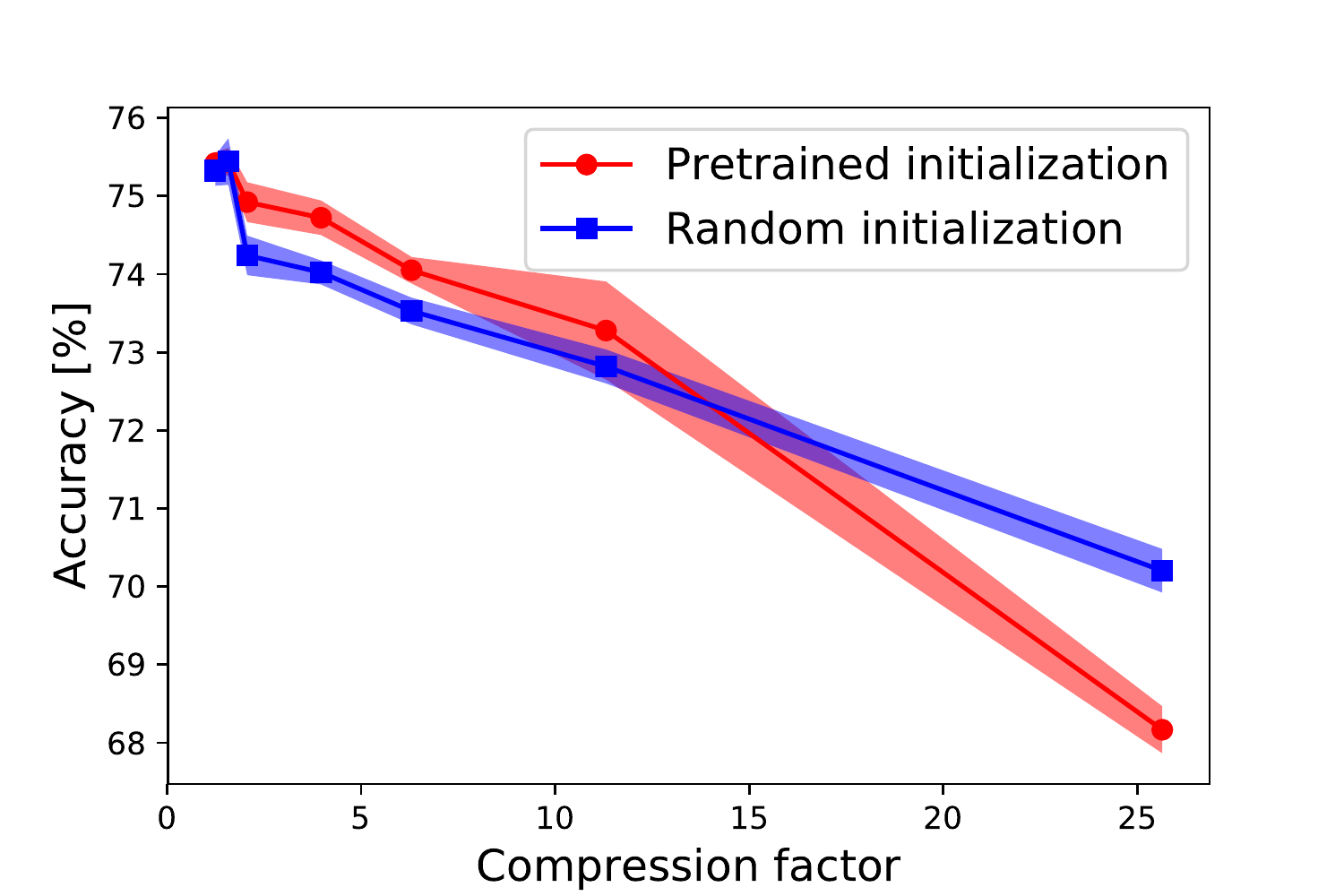}
	\caption{Random vs. pretrained initialization for fine-tuning. From left-to-right: C-NET, Resnet-20-16 and Resnet-20-64. Top row is Cifar10 and bottom row is Cifar100.}
	\label{fig:inits}
\end{figure}

Figure~\ref{fig:inits} shows the results of this experiment.
It can be noticed that irrespective of CF, random initialization are always as good or better than pretrained initialization.
Random initialization tends to train better under heavy pruning due to the $\ell_2$-regularized nature of the random initialization.
Prior works have made similar claims for unstructured pruning~\cite{zhou2019deconstructing, liu2018rethinking}.

\subsection{On the value of channel-wise pruning policies.}
Not all policies for deciding which channels to prune need an associated policy of how much to prune.
Rather, a direct policy for which channels to prune could implicitly decide how much to prune as well~\cite{liu2017learning, lebedev2016fast, zhou2016less}.
Some works even weigh the importance of channels before deciding which channels to prune~\cite{Luo_2017_ICCV, Yu_2018_CVPR}.
While we only consider methods that completely freeze pruned channels or remove them altogether, some methods allow soft-pruning at prune-time during which, pruned channels could be recovered back if they are needed~\cite{he2018soft, Wang_2018_ECCV}.
Also note that while we isolate this part of the policy for a controlled study.
Most methods use this as one of the stages in their overall strategy for pruning.
They use a hierarchical search space, which is as large as the one described in equation~\ref{eqn:all_profile}.
Typical channel-wise pruning policies use some metrics measured from the layer's tensors such as pruning the channels with the least expected $\ell_1$ of weights etc ~\cite{srinivas2015data, gomez2019learning, zhu2017prune}.
Molchanov's work used the first order equivalent of parameter saliency as described by Lecun \etal;
We refer to them simply as ``Taylor" features~\cite{molchanov2016pruning, lecun1990optimal}.
Note that saliency and the Taylor features are only applicable if we train and prune layer-wise or iteratively.
Among most SOTA, the component that is used to perform this decision is either using $\ell_1$ and Taylor~\cite{molchanov2016pruning, molchanov2017variational, ashok2018nn, he2018amc, zhuang2018discrimination, yu2019network}.
Therefore, we decided to isolate this component out of these SOTA and use them in a standalone fashion.
This enables to control for other factors.
In this section we consider three channel-wise policies that follow the setup of equation~\ref{eqn:channel_wise} but with $k_i = k,  \forall i$, that is retain the same percentage of channels in every layer.
Therefore, the job of the channel-wise pruning policy is to explicitly determine which channels to prune in a layer, given an estimate of how many channels are expected to be retained.

\begin{enumerate}[wide, labelwidth=!, labelindent=0pt, nosep,before=\leavevmode\vspace*{-1\baselineskip}]
	\item \textit{Random:} This strategy is the same as the policy $\pi_r$ described in equation~\ref{eqn:random_channel}.
	\item $\ell_1$: In this strategy, we sort all channels in order of their $\ell_1$ and retain channels from the top. 
	The featurization can be defined as $\Phi_{\ell_1}(L_t, w_t) = \{ \vert w_t^1\vert, \dots  \vert w_t^{c_t} \vert \}$, for a layer $L_t$.
	We consider retaining only $kc_t$ channels in layer $t$,  our policy is therefore to choose the $kc_t$ channels that have the top $\Phi_{\ell_1}(L_t, w_t)$ values. 	
	\item \textit{Taylor}: This featurization was originally described in the work by Molchanov \etal and is the center-piece of their pruning strategy~\cite{molchanov2016pruning}.
	We define the metric here as follows:
	For layer $L_t$ with $c_t$ channels the metric corresponding to channel $i$ is, $ \Phi_T(L_t^i,w_t) = \expectation_{k_h, k_w, c_{t-1}} \Big \vert  \frac{\partial \epsilon} {\partial L_t^i} L_t^i \Big\vert$,
	where, $\epsilon$ is the validation error of the entire network and the expectation is over the non-channel dimensions of the layer's tensor~\cite{lecun1990optimal}. 
	The final metric is averaged over all the data samples.
	Consequently, $\Phi_T(L_t,w_t) = \{ \Phi_T(L_t^1,w_t), \dots \Phi_T(L_t^{c_t},w_t) \}$ can be used as a metric in a manner similar to $\ell_1$. 
	Similar to $\ell_1$ above, the pruning policy is to choose the $kc_t$ channels that have the top $\Phi_{T}(L_i,w_t)$ values.
	While it is possible to use the $\ell_1$ feature without training the network layer-wise after pruning every layer, the same is not possible for the Taylor features.
	Therefore we prune layer-wise and fine-tune after each layer-wise pruning to enable fair comparison.
\end{enumerate}

\begin{figure*}[t]
	\includegraphics[width=0.325\linewidth]{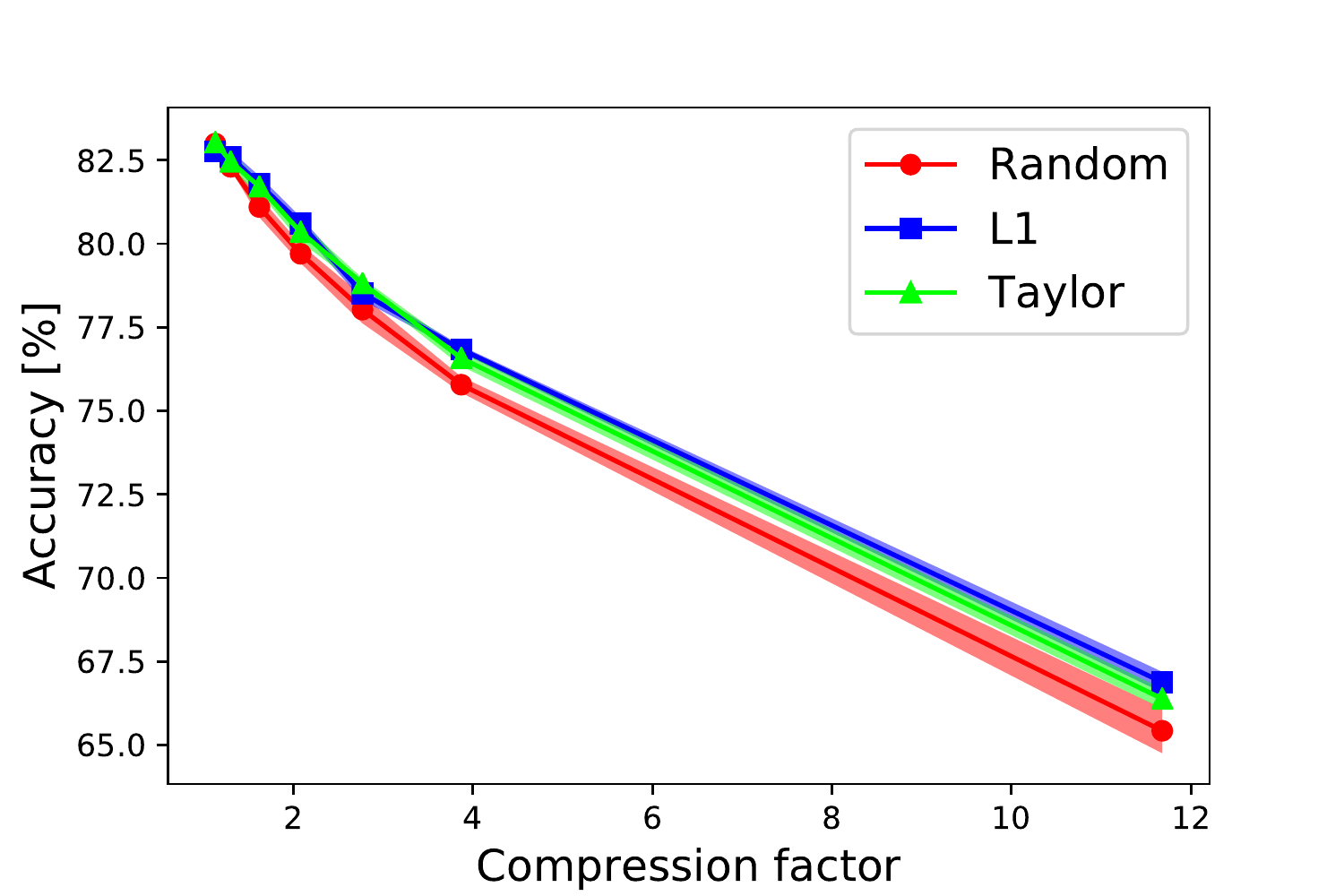}
	\includegraphics[width=0.325\linewidth]{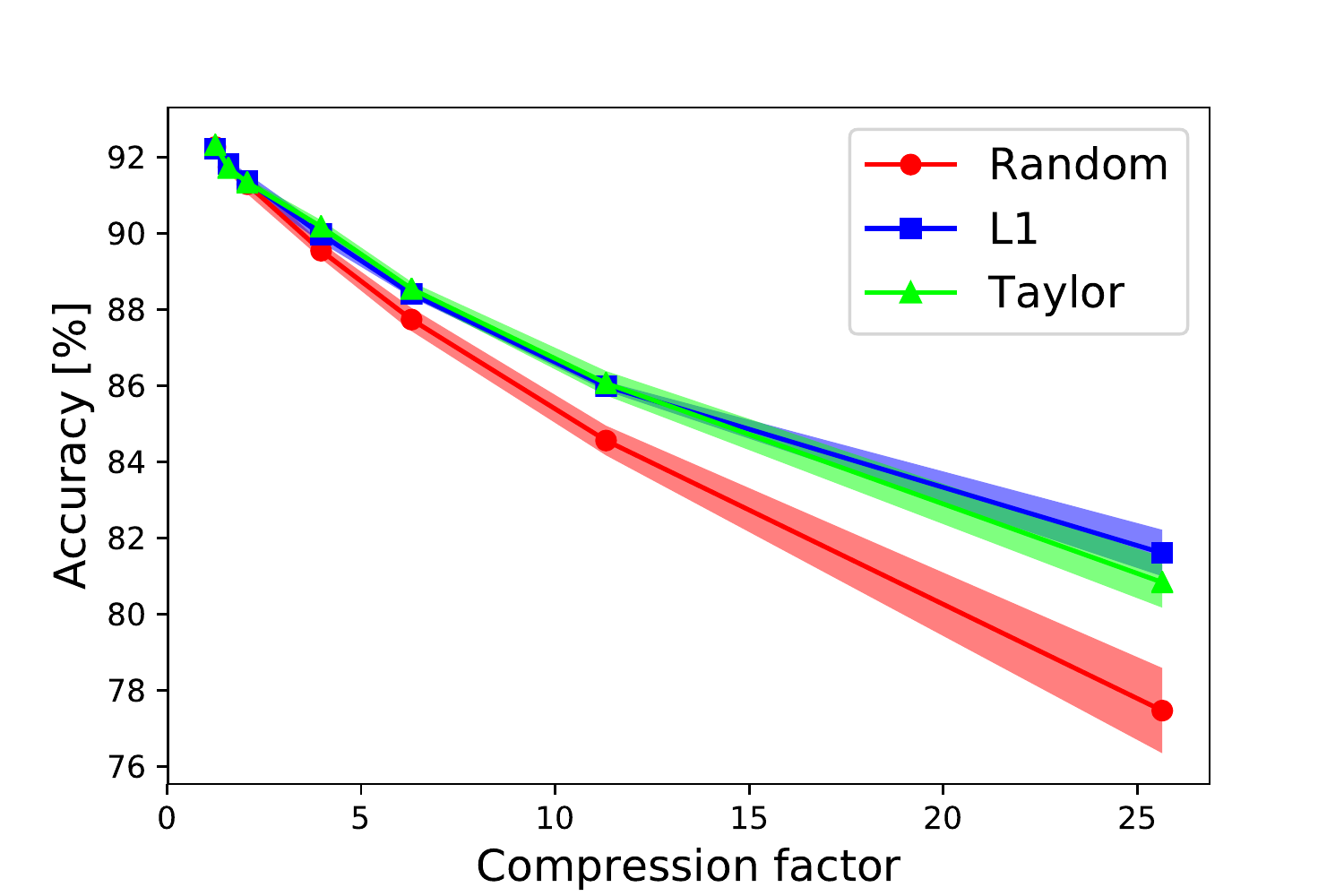}
	\includegraphics[width=0.325\linewidth]{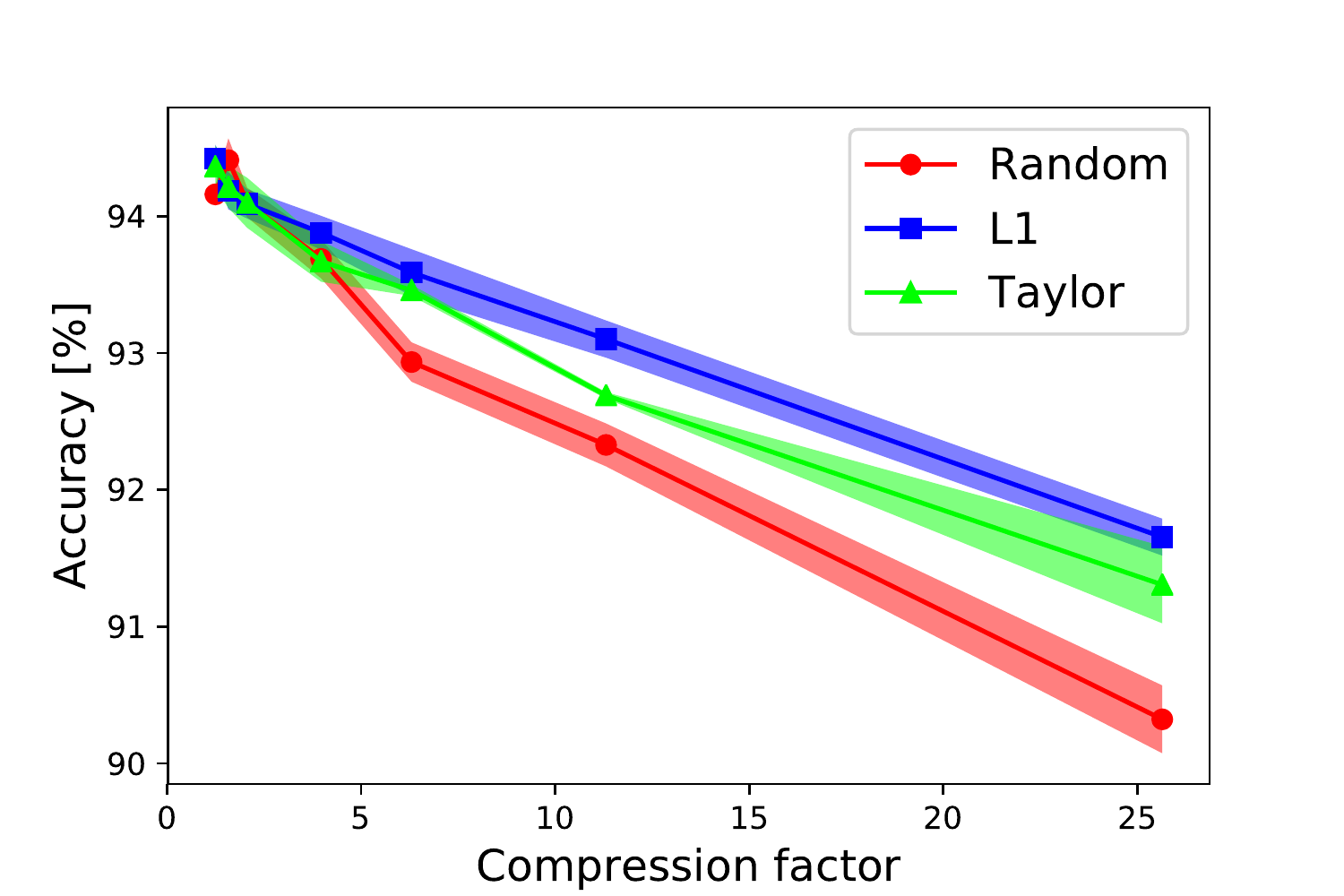}
	\includegraphics[width=0.325\linewidth]{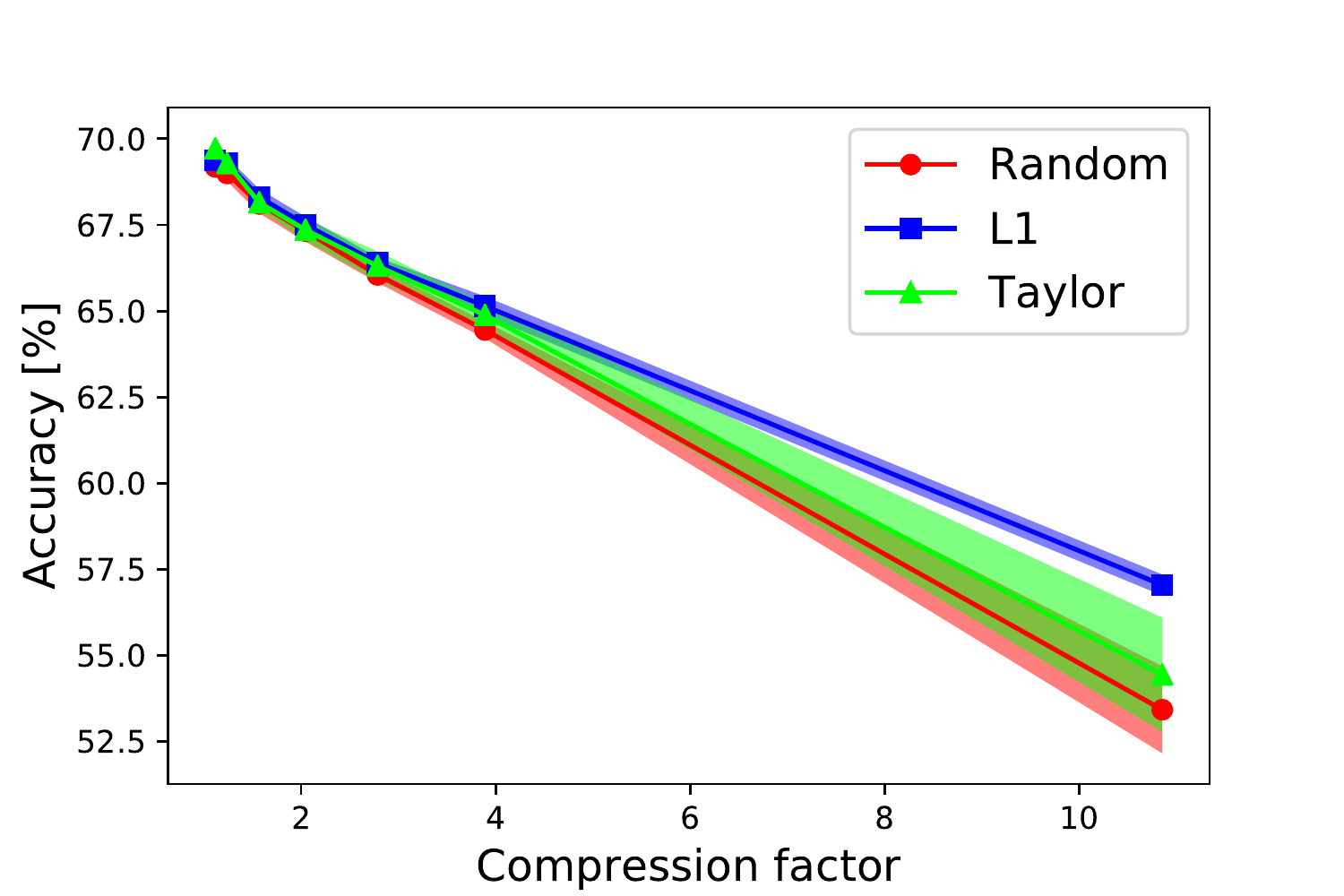}
	\includegraphics[width=0.325\linewidth]{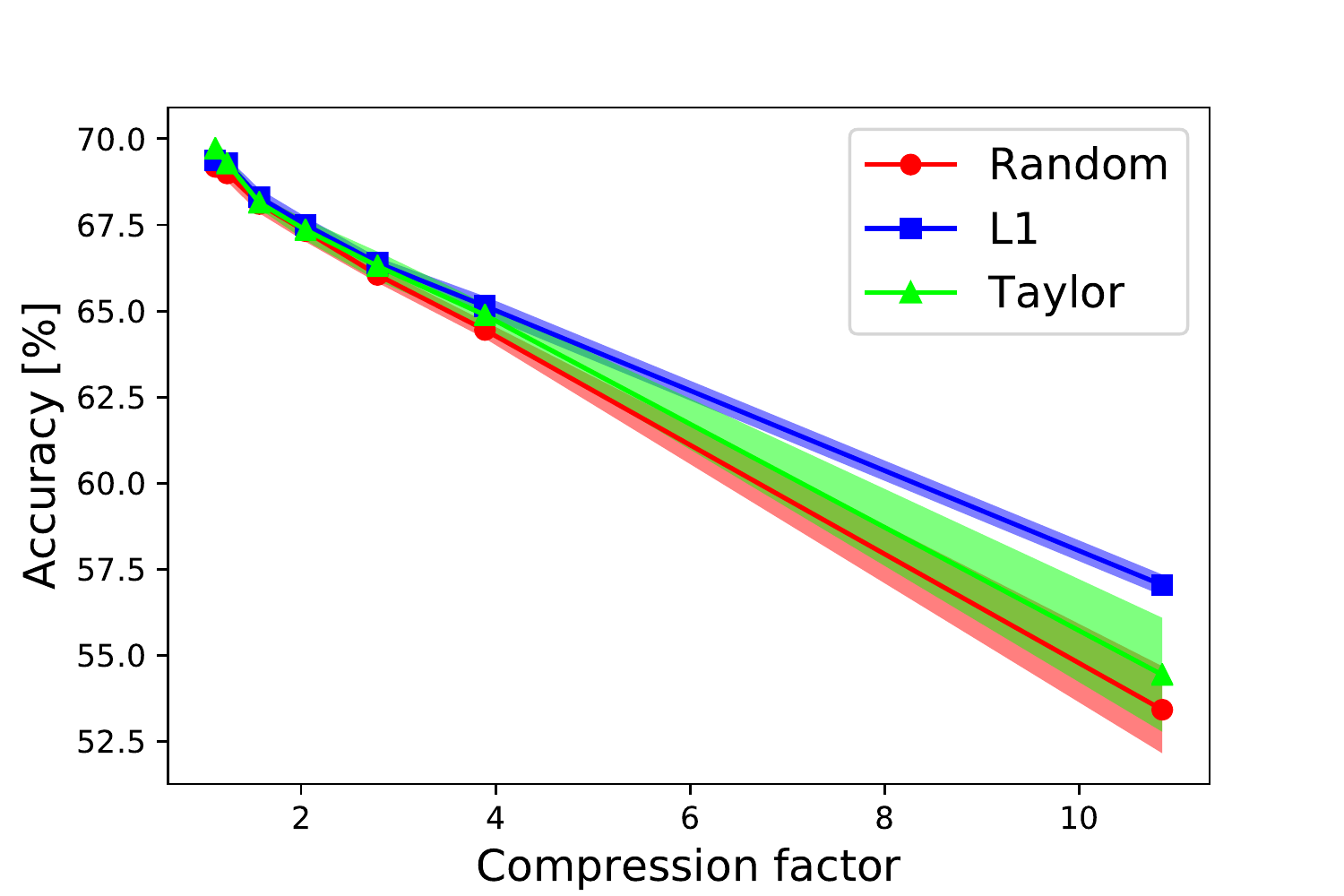}
	\includegraphics[width=0.325\linewidth]{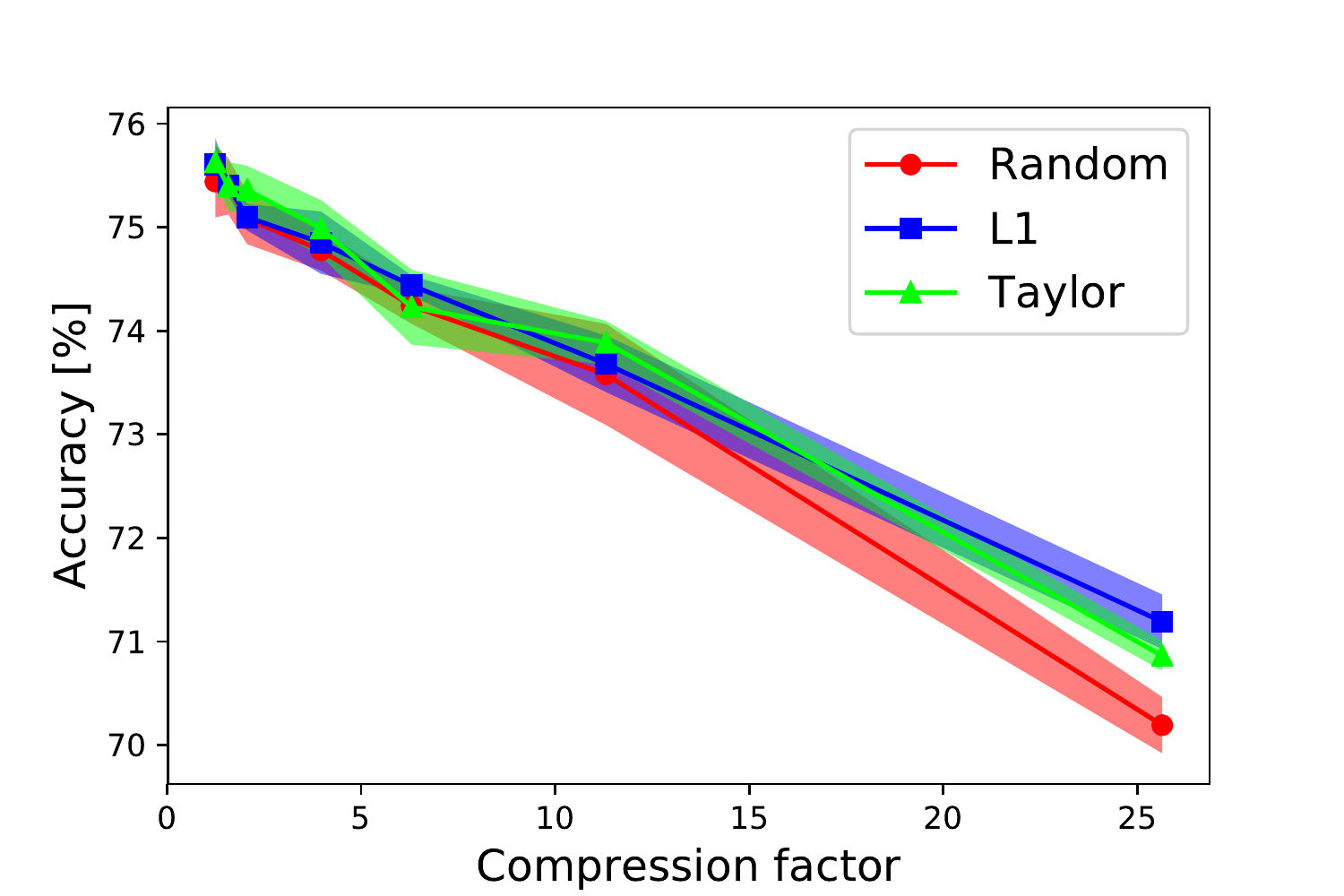}
	\caption{Which channels to prune? From left-to-right: C-NET, Resnet-20-16 and Resnet-20-64. Top row is Cifar10 and bottom row is Cifar100.}
	\label{fig:channel}
\end{figure*}

Figure~\ref{fig:channel} shows the effect of these metrics on the decisions for which channels to prune.
We do observe noticeable differences in some cases, particularly under heavy compression, Taylor and $\ell_1$ metrics have some advantages.
Random choice for channel removal seem to be in general as good as $\ell_1$, particularly for lower compression factors. 
However, $\ell_1$ seem to be the best performing metric across all compression factors.
While Mittal \etal show with empirical evidence using similar experimental setup that for unstructured pruning, randomly removing weights seem to be as good as $\ell_1$ and other such metrics, we show that for structured pruning, this might not necessarily be the case~\cite{mittal2018recovering}.
Given that in the regions of most pragmatic compression factors with least loss of accuracy, all the methods yield similar performance, we concur with them that in most common cases, random is just as good.
Therefore, this result also serves as an independent validation of part of their arguments and extends it to the channel pruning context.
We use this insight  in our RL-based search (described in section ~\ref{sec:case}), to learn a policy of how many channels to prune, while using the random strategy for which channels to prune.

\noindent\textbf{Key insights from section \ref{sec:taxonomy}:} For decisions over initialization and selection of channels to prune, random choices perform as good as common baselines for most cases.
Given that random channel-selection policies seem to be able to recover the performance, we continue with the study of layer-wise pruning policies.

\section{Layer-wise Pruning Profiles}
\label{sec:layerwise}
While in the previous section, we analyzed the options for deciding which channels need to be retained, we assumed that for every layer, we have a fixed heuristic $k$ for how many channels are to be retained.
In this section, we will study policies that decide how many channels are to be retained as well. 
Several types of policies can be used to decide how much to prune per-layer.
These can range from heuristics such as a equally-distributed pruning policy across all layers, to learning a policy either online or offline. 
The policies can be learned either using an RL or gradient-based methods, on observing some properties of the network or layer, such as the correlation analysis of layers~\cite{xavier}.
In iterative pruning techniques such as the work by Molchanov \etal the profiles are also progressively adapted~\cite{molchanov2016pruning, lecun1990optimal}.
On the other hand some techniques use an RL agent to directly learn the policy~\cite{yu2019universally, yu2019network, ashok2018nn, he2018amc}.

Thus far we have considered policies $\pi$ where the value $k$ of how much to prune every layer is heuristically fixed.
Here, we consider four general categories of layer-wise pruning policies that determine how much to prune per-layer and making the second-level decision of which channels to prune using a random strategy such as equation{~\ref{eqn:layer_wise}.
	
	\begin{enumerate}[wide, labelwidth=!, labelindent=0pt, nosep,before=\leavevmode\vspace*{-1\baselineskip}]
		\item \textit{Equally-distributed profiles:}
		Considering $l$ layers, $p_e = \{k\}_{\times l}, k \in [0,1]$.
		\item \textit{Increasing profiles}: Some prior online learning works argue that a profile that is monotonically increasing, with less pruning on initial layers and more pruning on final layers is emergent out of their profile search~\cite{singh2019darc, ashouri2018fast}. 
		In this work we construct increasing profiles $p_i=\{\frac{si}{l}\}, i\in \{1, \dots l\}, s\in[0,1]$ of several slope $s$ resulting in several compression factors.
		\item \textit{Decreasing profile}: Since we consider equally-distributed and increasing profiles, we also use symmetric decreasing profiles for a complete study.
		\item \textit{Random profiles}: We created random profiles for a network with $l$ layers as, $p_r \sim \{\mathcal{U}(0.3, 0.9)\}_{\times l}$.
		This implies that for any layer, we randomly sampled (using a uniform distribution) a value of how much to prune between $0.3$ implying that we retain only $30\%$ of the channels, to $0.9$ implying that we retain $90\%$ of the channels.
		We chose a non-zero lower-bound because at closer-to-zero lower-bound the pruned network ran into instabilities.
	\end{enumerate}
	
	\noindent\textit{Reasoning for choice of our baselines:} Given that we sample random profiles exhaustively from the entire search space, every SOTA method's solution is expected to lie within the space of this search space as well.
	While random/exhaustive search is not an efficient search strategy it is the most effective at finding the top profiles as will be established later section.
	
	\begin{figure*}[t]
		\includegraphics[width=0.325\linewidth]{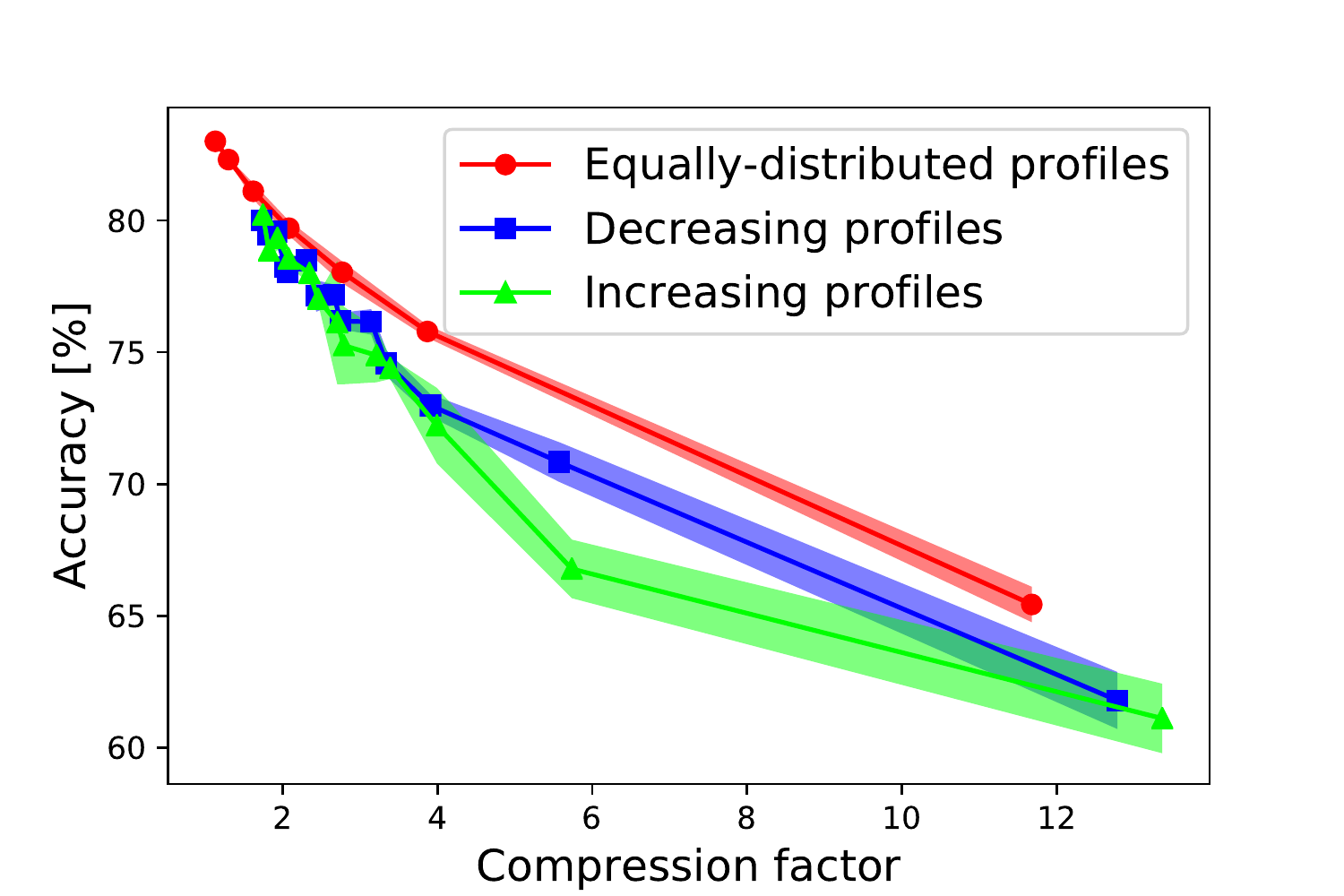}
		\includegraphics[width=0.325\linewidth]{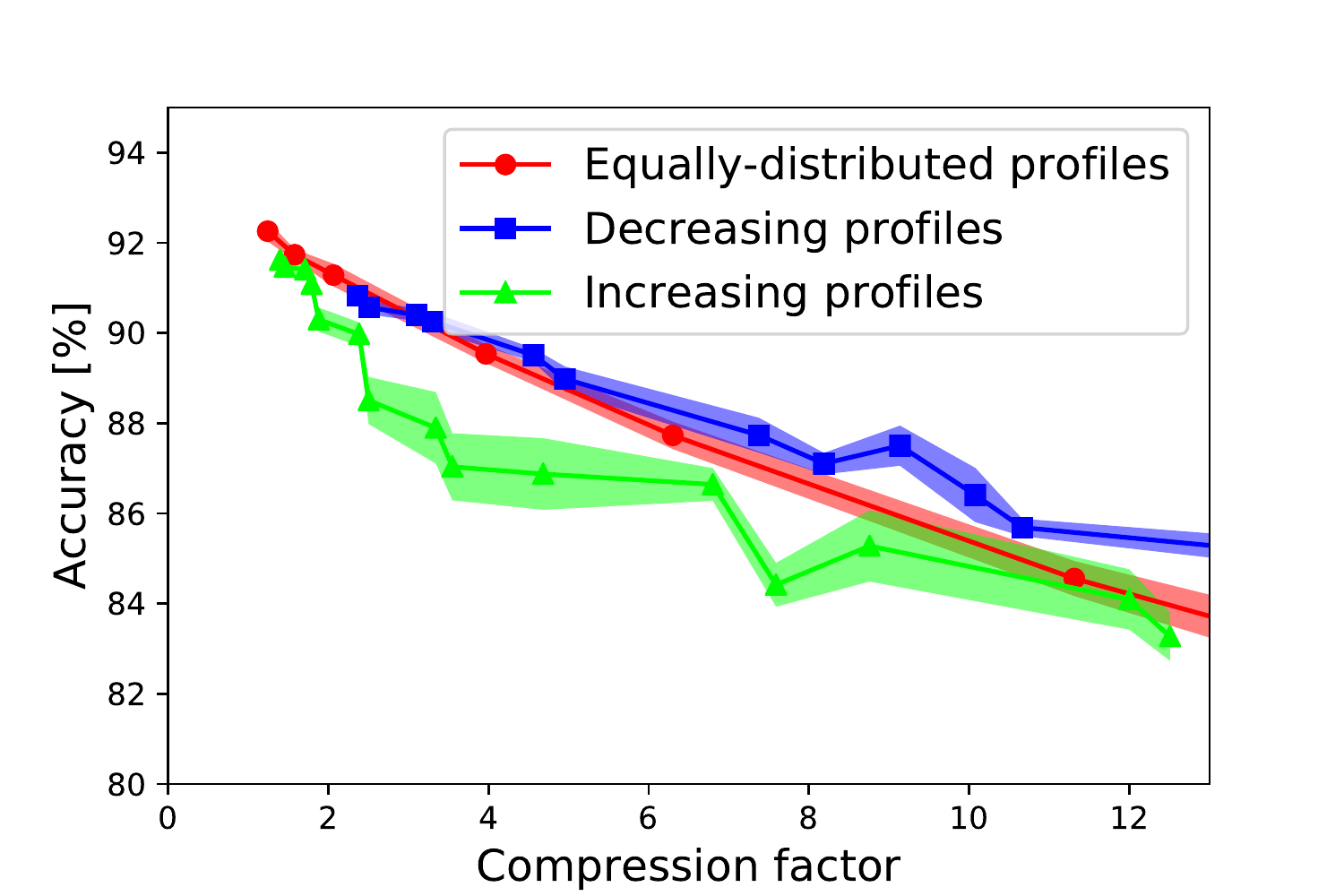}
		\includegraphics[width=0.325\linewidth]{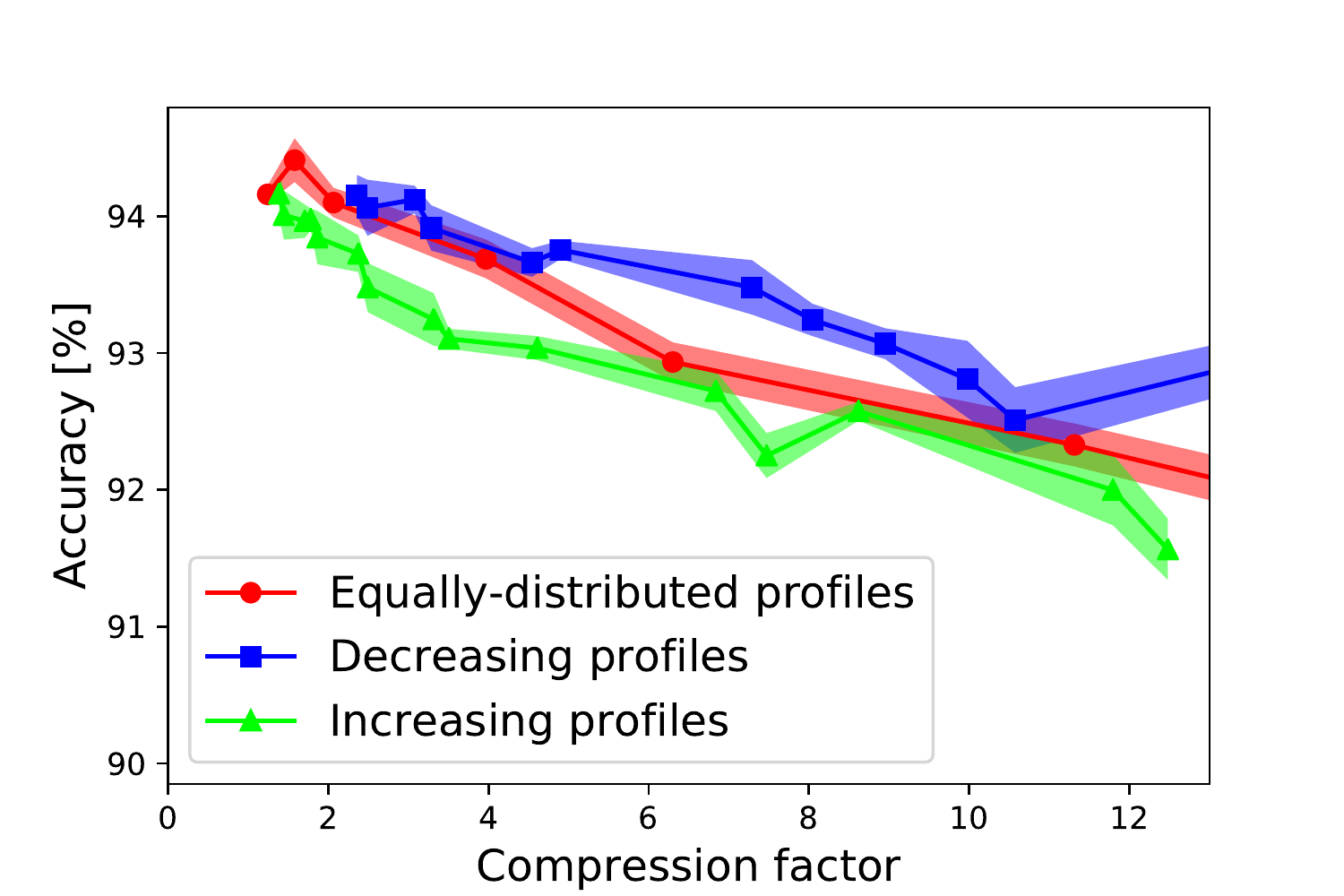}
		\includegraphics[width=0.325\linewidth]{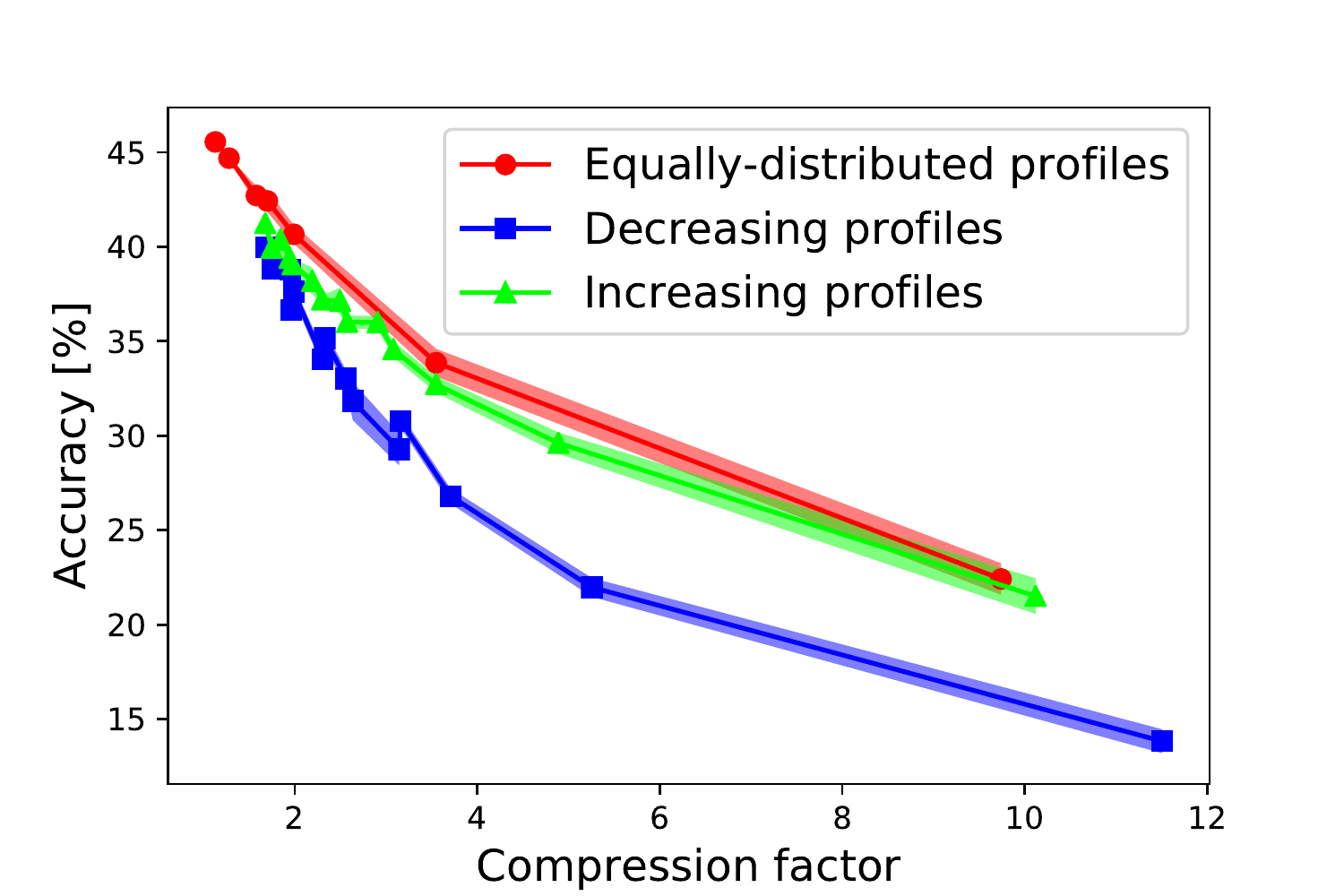}
		\includegraphics[width=0.325\linewidth]{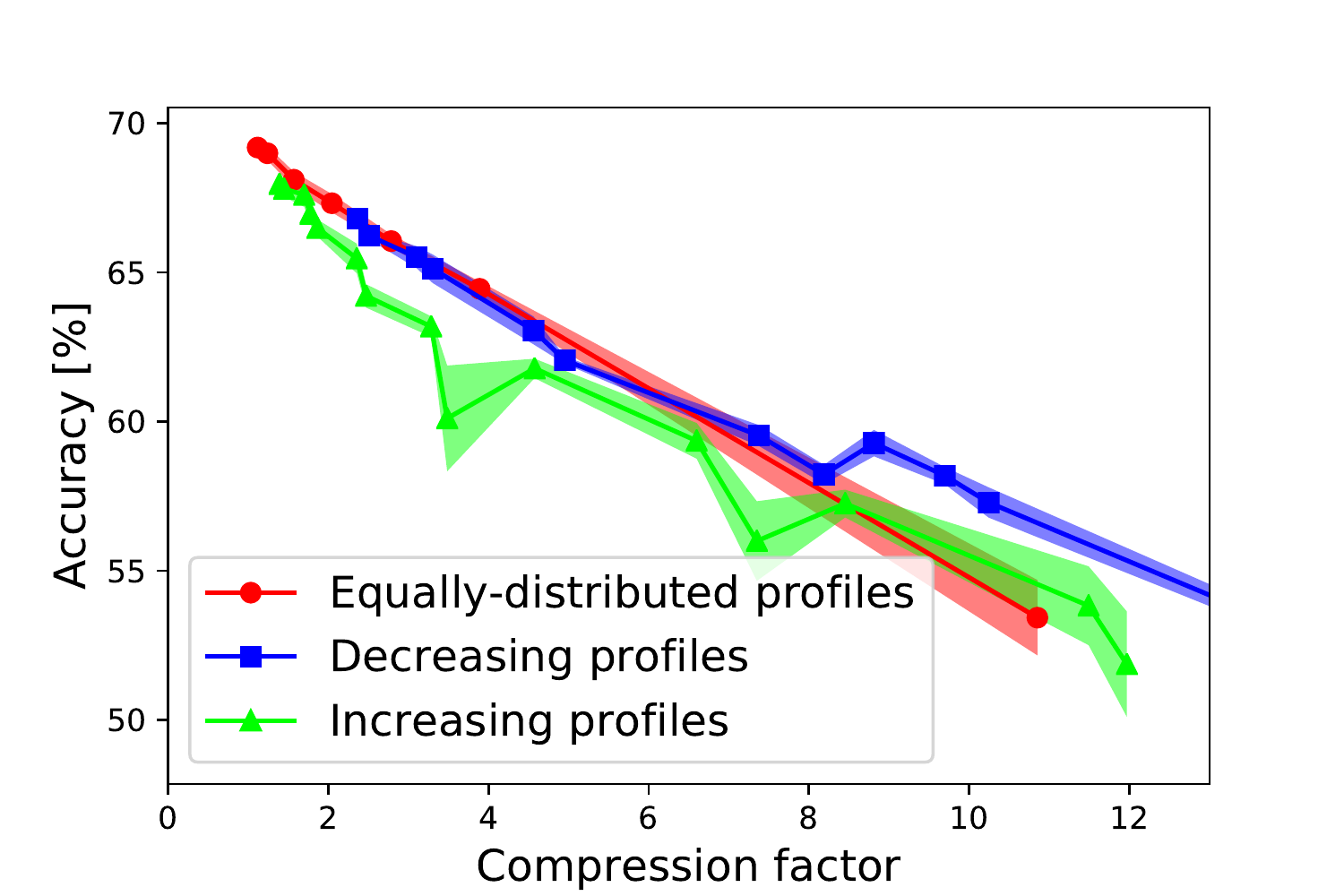}
		\includegraphics[width=0.325\linewidth]{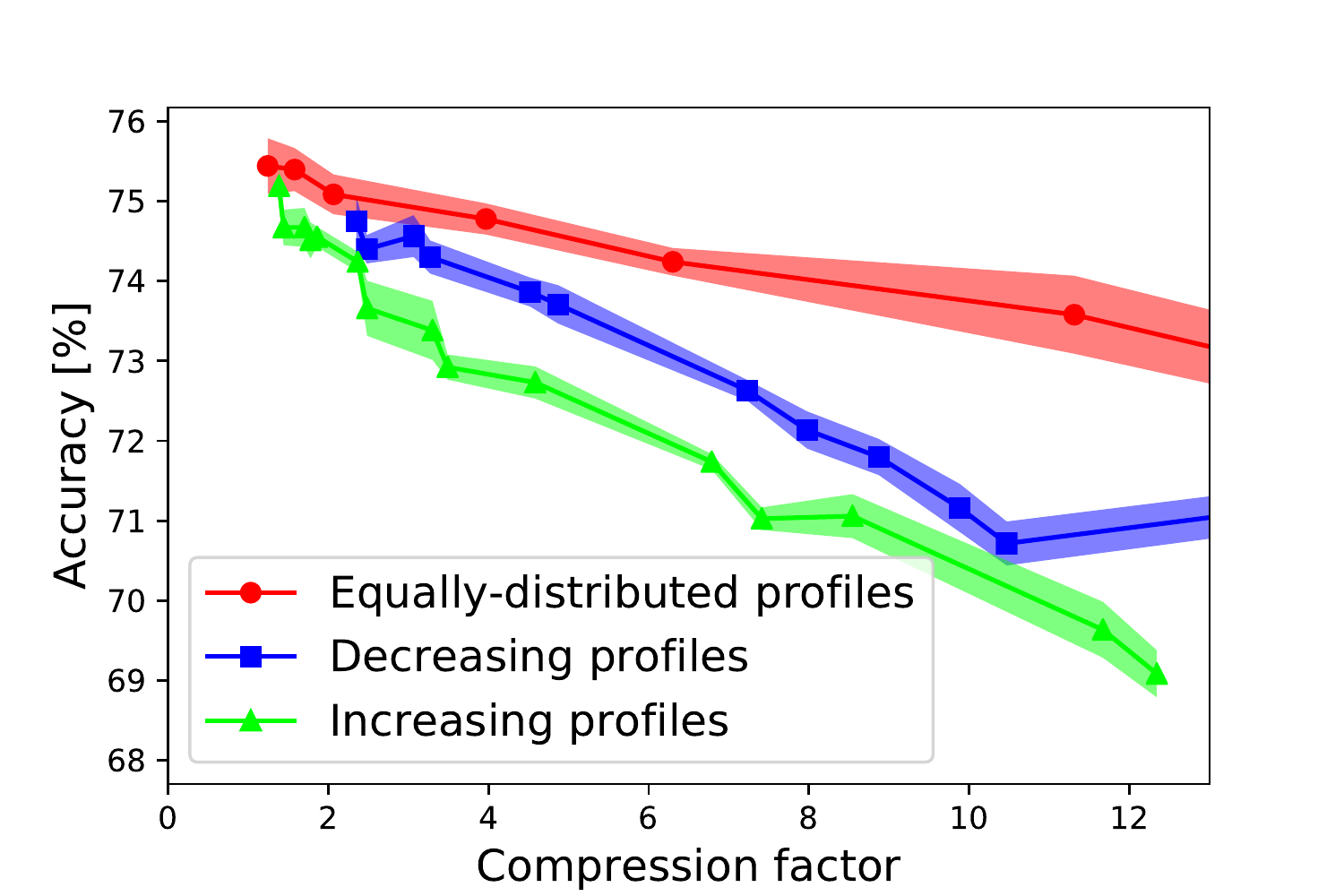}
		\caption{How much to prune? From left-to-right: C-NET, Resnet-20-16 and Resnet-20-64. Top row is Cifar10 and bottom row is Cifar100}
		\label{fig:howmuch}
	\end{figure*}
	
	Figure~\ref{fig:howmuch} demonstrates that there are no significant differences between profiles for most compression factors though decreasing profile performs worse than equally-distributed for Resnets.
	There is however the question of if there are profiles that can be significantly better than hand-crafted heuristic profiles that are used in figure~\ref{fig:howmuch}.
	There are some works such as those by Ashok \etal and He \etal that even try to use reinforcement learning to find such profiles~\cite{ashok2018nn, he2018amc} for a given network.
	In the next section, we propose one of our own RL approach, which, unlike those mentioned above, does not have to search a new profile for the target network.
	
	We sampled about $1200$ random profiles in each experiment ($600$ for TinyImagenet due to cost-constraints), in the hope of finding ones that are closer to the base model's accuracies, with the caveat that since the search space is monumental, we might not be able to find the best possible profiles.
	We limit the search space to a compression factor of 6x.
	While this search was not objective-driven and one may argue that there could be profiles that are significantly better, we could use this as a reasonable approximation of the search space of the policy-based search algorithm proposed in section~\ref{sec:case}. 
	Also note that these profiles are found using four Cifar10/Cifar100 trained networks and when we pick the top profiles, we measure and plot them against a fifth unseen Cifar10/100 trained network, out-of-the-box without optimizing the profiles.
	
	\begin{figure*}[t]
		\includegraphics[width=0.32\linewidth]{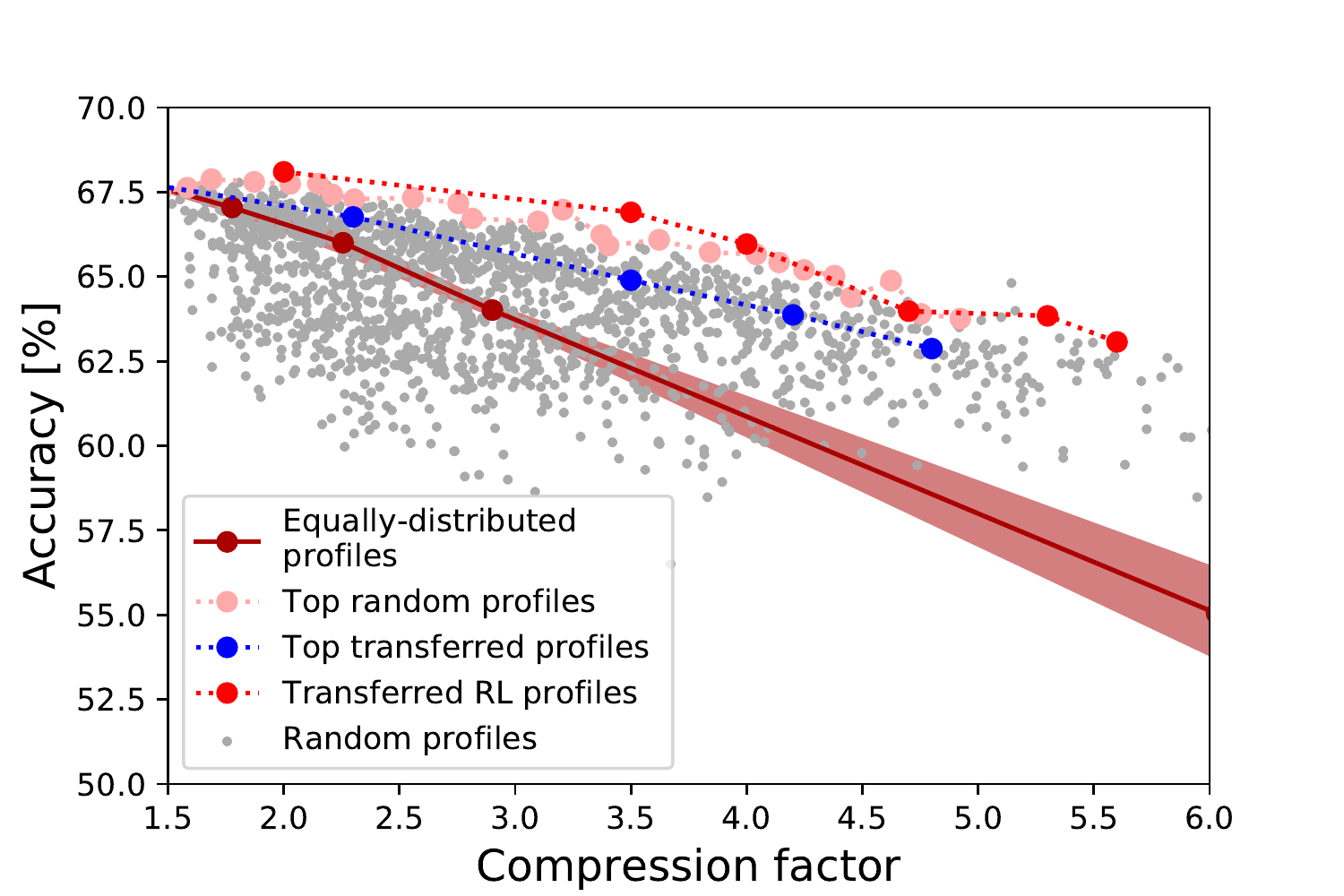}
		\includegraphics[width=0.32\linewidth]{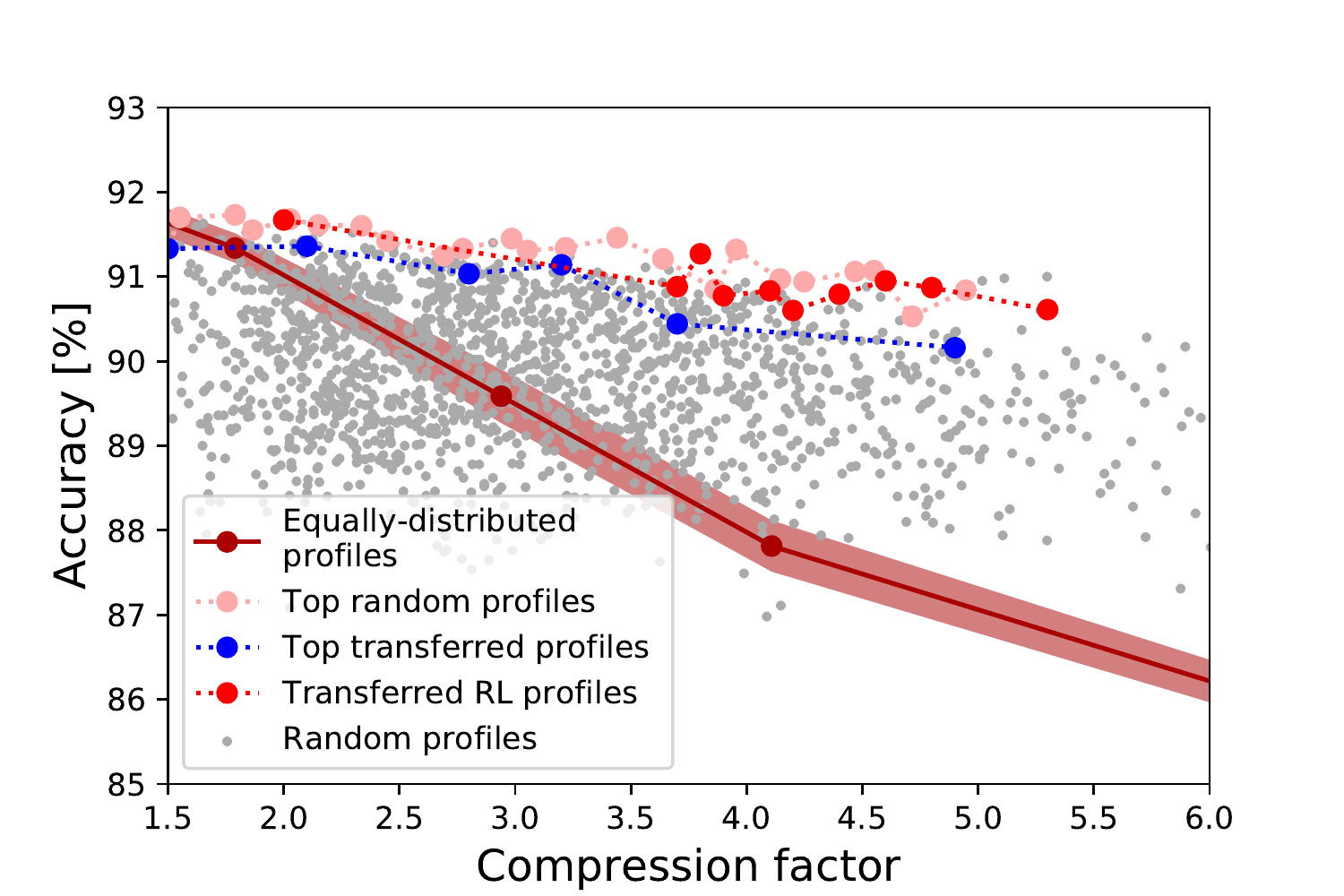}
		\includegraphics[width=0.32\linewidth]{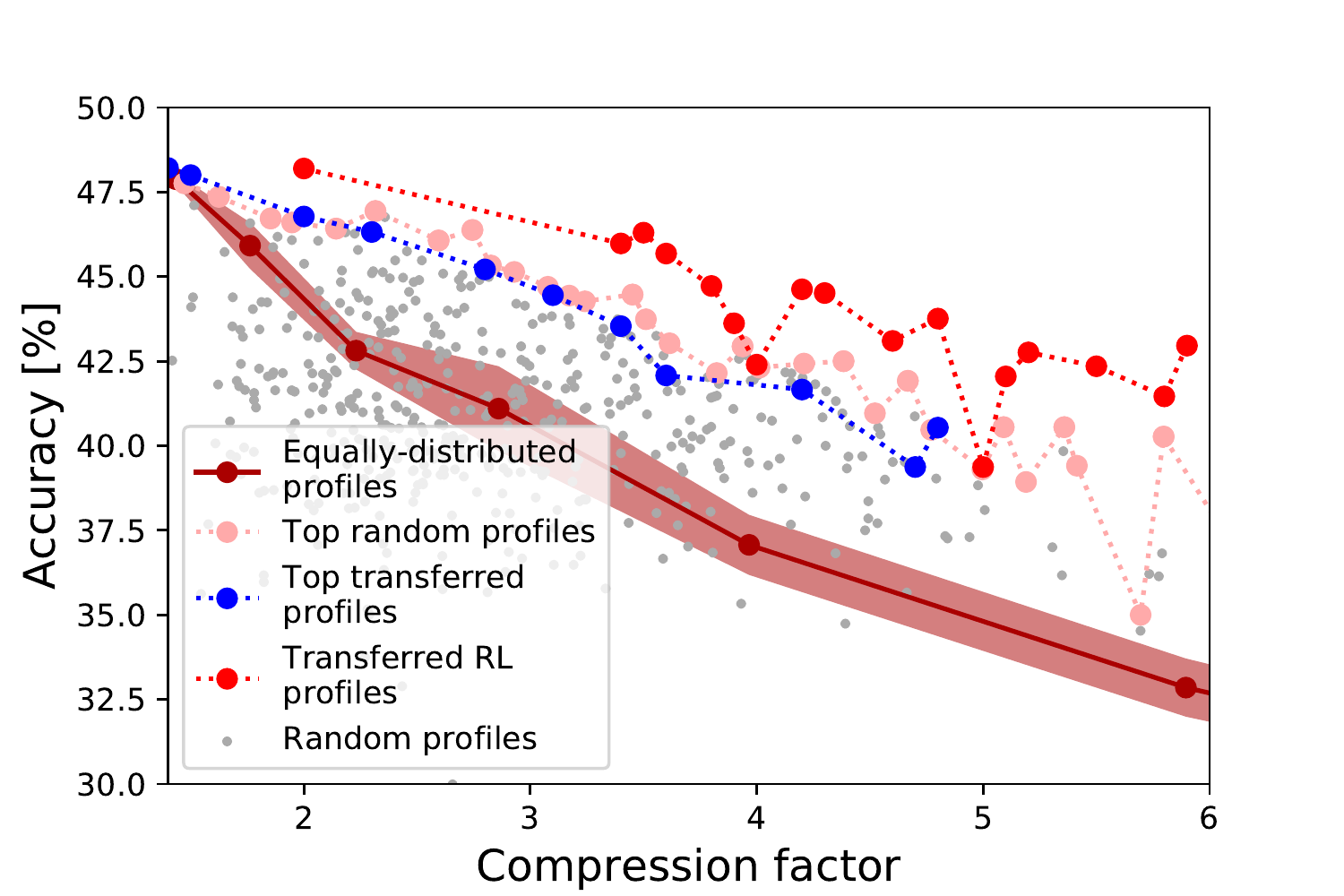}	
		\caption{Random profiles and transferred profiles compared against the equally-distributed profiles for Resnet-20-16. The graph on the left is for the Cifar100 dataset with the transferred profiles coming from the best Cifar10 profiles that are highlighted in pink on the graph in the center. The center graph is for the Cifar10 dataset with the transferred profiles coming from the best Cifar100 profiles highlighted in pink on the graph in left. The graph on the right is the best Cifar10 and Cifar100 transferred to TinyImagenet dataset.}
		\label{fig:rand-profile}
	\end{figure*}
	
	Figure~\ref{fig:rand-profile} shows the random profiles that we have searched.  
	As can be seen, at all CFs, there exists profiles that perform close to the base model performance. 
	Crucially, pruning profiles are very important since only a few profiles at every CF are at the top while a majority of the profiles lose performance.
	This suggests that such a random-search is a strong baseline.
	
	\subsection{Transferability of Pruning Profiles}  
	
	Exploring a random search space for each base model is not a scalable solution.
	It is expensive to search for and find a pruning profile for every CNN trained, as the search algorithms often require an order of magnitude more compute, power and time than pruning and are not sustainable. 
	In the previous sections, we have argued for reduced emphasis on all dataset-related statistics for pruning policies and emphasized on the profile, while all else being decided by random.
	An apparent follow-up hypothesis is if the profiles are truly tied to the architecture and not to the dataset, even though we discover these profiles from trained models.
	To wit: Will we be able to discover a top profile from a network on one dataset, and simply use the profile to prune another network trained on a completely different dataset and expect it to still be satisfactory?
	We test this hypothesis by conducting a pruning profile search on one dataset and using the best profile at many CFs on another dataset.
	
	In figure~\ref{fig:rand-profile} along with the randomly trained profiles, we also highlight the best profiles discovered (in pink) from our random search at each CF.
	We make a random selection of these profiles spaced throughout the CFs, and transfer them out-of-the-box without optimizing them further, to the other dataset and measure the performance across the second unseen dataset (in blue).
	It can be noticed that the best profiles discovered from one of the dataset are also among the best profiles, if only slightly lower in performance, in the search space of the other dataset at all levels of compression.
	Taking this idea further, we find in figure~\ref{fig:rand-profile} that the best profiles from both Cifar10 and Cifar100 also transfer out-of-the-box to TinyImagenet~\cite{tiny}.
	This is a strong result considering the chances an arbitrary pruning profile from a different dataset to be among the top profiles discovered from the considered dataset. 
	This generalization of pruning profiles can help build a scalable channel pruner.
	
	\noindent\textbf{Key insights from section \ref{sec:layerwise}:} In summary, we can draw the following insights from this section. 
	\begin{enumerate*} 
		\item When it comes to channel pruning, what matters most are the layer-wise pruning profiles. 
		Plasticity of neural networks during fine-tuning can re-capture the performance even if most other stages are left random.
		\item There exists certain layer-wise profiles that are in general tied to particular architectures that even when trained from scratch on a new dataset, which had no relationship to the discovery of said pruned architecture, performs as good as or close to the best channel pruning profile that could be discovered with the same level of compression using the said dataset.
	\end{enumerate*}
	
	While we have demonstrated that the top pruning profiles are transferable to unseen target datasets, we further our investigation in two directions. 
	Firstly, we would like to try and better the transferred profiles and perform close to the top profiles on the target datasets.
	Secondly, we would like to be able to find these profiles using a strategy that is better than exhaustive random search.
	To this end, we develop a novel pruning profile searching system using RL in the subsequent section.

\section{Reinforcement Learning Framework}
\label{sec:case}
Reinforcement learning (RL) develops policies for sequential decision-making problems. 
Its fundamental principle is that an agent makes decisions based on the observation received, and tries to maximize a reward signal.
Recently, RL has emerged as a powerful and general approach in various domains from complex games \cite{silver2016mastering, tian2017elf, berner2019dota, vinyals2019alphastar} to simulated robotics tasks \cite{rajeswaran2017learning, andrychowicz2020learning, hwangbo2019learning}, all the way to neural architecture search~\cite{zoph2016neural, pham2018efficient, cai2018proxylessnas}.
As was discussed before, RL-based methods such as N2N and AMC performs channel-pruning by learning layer-wise policies for channel pruning, albeit they learn a new policy for every network individually~\cite{ashok2018nn, he2018amc}. This requires significant searching for every network.
Other works also perform layer-wise unstructured sparsity pruning similarly~\cite{lillicrap2015continuous}.

In this section we propose an RL-based method that uses random strategy for deciding which channels to prune and learns a policy to produce out-of-the-box transferable layer-wise pruning profiles similar to the formulation in equation~\ref{eqn:layer_wise}.

\subsection{Problem Formulation}
We train a RL policy to produce layer-wise pruning profiles 
\begin{enumerate*}
	\item of the form described in equation~\ref{eqn:layer_wise} (produce the $\beta$s) while also constraining the overall network CF around an expected CF.
	\item that can transfer to other datasets and out-of-the-box be close to the top profiles that could be discovered using the transferred dataset.
\end{enumerate*}
\begin{figure}[t]
	\includegraphics[width=\linewidth]{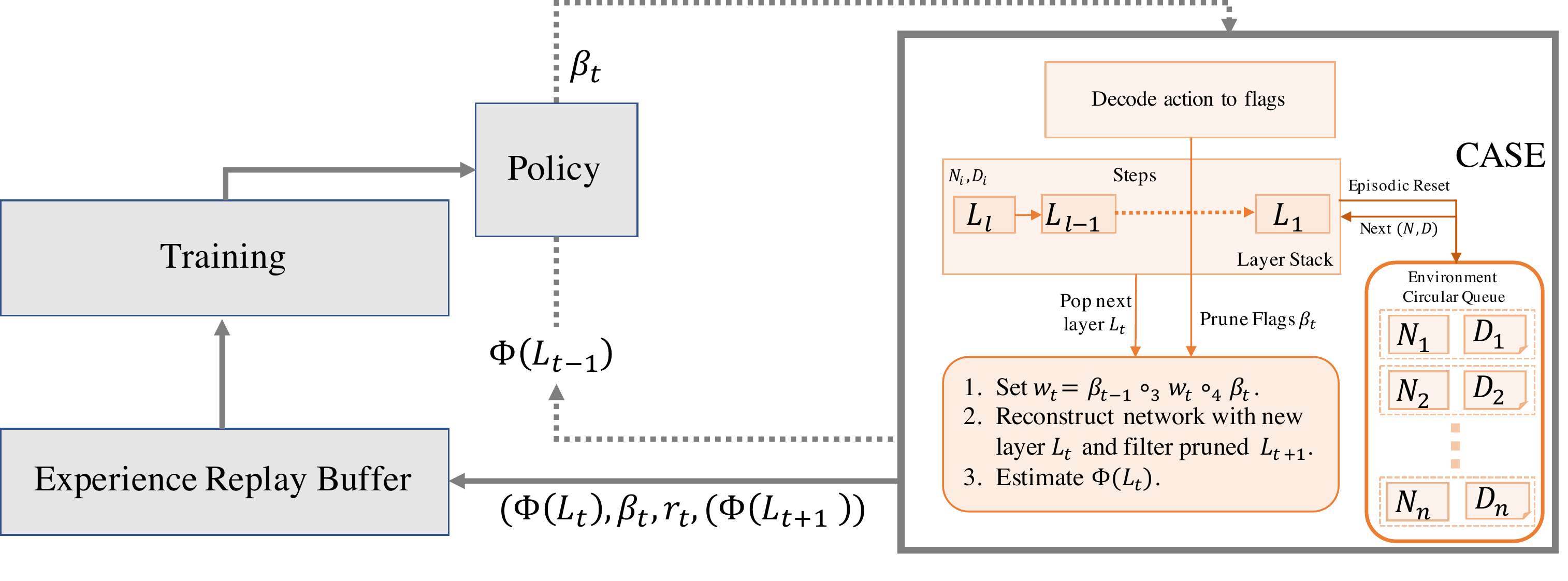}
	\caption{Illustration of our training.}
	\label{fig:case}
\end{figure}

We propose a Markov Decision Process (MDP) formulation to achieve this.
Figure~\ref{fig:case} illustrates our setup\footnote{Detailed working is discussed in the supplementary.}.
The agent generates pruning decision $\beta_t$ for layer $L_t$ after observing the features of the layer $\Phi(L_t, w_t)$.
The proposed MDP therefore operates layer-wise with each layer being a step.
The episode ends as the entire pruning profile is produced and a new network is sourced from the queue for the next episode.

\noindent\textbf{Action space:}
The action $\beta_t \in [0,1]$ for the $t^{\text{th}}$ step for layer $L_t$ leads to retaining $\beta_tc_t$ channels being retained.
For each channel in this layer we use random strategy as described in equation~\ref{eqn:layer_wise} to decide which channels to remove.
At the end of the episode, the agent will produced a layer-wise pruning profile.

\noindent\textbf{Observation space:}
We repurpose the Taylor metric as observations~\cite{molchanov2016pruning}.
For layer $L_t$ with $c_t$ channels, and a pruning action $\beta_t \in [0,1]$ the observations corresponding to channel $j$ are,
\begin{equation}
\Phi(L_t^j \vert \beta_t) =  \begin{cases} 
\expectation_{k_h, k_w, c_{t-1}} \Big \vert  \frac{\partial \epsilon} {\partial L_t^j} L_t^j \Big\vert  & \text{if } \Bernoulli(\beta_t)= 1 \\
0 & \text{otherwise}
\end{cases},
\label{eqn:observations}
\end{equation}
where, $\epsilon$ is the validation error of the entire network.
The observation of the entire step is $\Phi(L_t) = \{\Phi(L_t^1) \dots \Phi(L_t^{c_t}), d_t\}$, where $d_t$ is a set of layer descriptions similar to the ones used by He \etal \cite{he2018amc}.
These features are derived from a direct consequence of change in the loss with respect to the actions taken at the previous time step.
Fore more details, refer the supplementary.

\noindent\textbf{Rewards:}
We use independent measurements of compression and accuracy and combine them into a novel conjoined reward term. 
More details on rewards are available in the supplementary.
The reward for the step $t$ of a $l$-step (number of layers in the network is $l$) episode is,
\begin{equation}
r_t =  \begin{cases}
0 & \text{if} \ t \neq l \\
\frac{A}{A_e} e^{-\frac{(C-C_e)^2}{2\sigma^2}}	& \text{otherwise},
\end{cases}.
\label{eqn:reward}
\end{equation}
where $C_e$ is an expected compression factor that is set at the environment-level, which is the compression at which we expect our profiles, $A_e$ is the expected accuracy that we desire, which is also the accuracy of the original unpruned model, $A$ is the accuracy that is produced by the model at the end of the episode, $C$ is the compression factor at the end of the episode and $\sigma$ is the severity of missing the expected compression.
Therefore, our rewards encourage the agents to maintain as high an accuracy as possible while requiring the compression factor to remain around the expected compression factor.
While using this reward we notice that, most RL agents learn to to fixate around the expected compression early in the training and try to search profiles around that CF, in order to maximize the accuracy.
Notice that the accuracy reward term can be greater than $1$ as well, for cases where the pruned model outperforms the unpruned model since compression can sometimes regularize models to generalize well.

\subsection{Learning}
To learn the policy, we use a circular queue of MDPs containing many base networks and datasets with which the network were trained $(N_i, D_i)$. At the beginning of a new episode a new $(N,D)$ pair is chosen. 
The episode begins at layer $L_1$ and ends at layer $L_l$, for a $l$-layer network.  
At the beginning of each episode, all $\beta_i = 1 ,i \in \{1, \dots l\}$.
At any time step $t, 1 \leq t \leq l$, the agent has produced actions $\beta_1, \dots \beta_t$ while $\beta_i = 1 $ for $i \in \{t+1, \dots l\}$.
We use this profile at every step to setup the network and produce the rewards appropriately.

The objective is to maximize the expected cumulative reward over several cycles of the circular queue. 
We use PPO, a state-of-the-art policy gradient algorithm that utilizes a policy network and a value network~\cite{schulman2017proximal}.
During training, the policy network interacts with the channel pruning environment and generates Gaussian-distributed actions given the features. 
In this case, the actions are recorded directly as $\beta$.
The value network estimates the expected cumulative discounted reward using the generalized advantage algorithm \cite{schulman2015high}.
We use RLLib's PPO implementation for our system~\cite{rllib}.

\subsection{Results}
We trained the policy using a queue of eight different Resnet-20-16 models, four trained using Cifar10 and four trained using Cifar100. 
We trained the policies with expected compression at 2x, 4x and 6x. 
We use the profiles that the policy has produced for the queue of models and out-of-the-box transfer them to other unseen Cifar10 and Cifar100 models.
We have plotted the results in figure~\ref{fig:rand-profile}.
It can be noticed that the profiles discovered using a targeted search perform significantly better than the transferred profiles searched from random search before.
The transferred profiles from RL search are closer to the best profiles discovered using the target dataset itself, which is significant.

\begin{figure*}[t]	
	\includegraphics[width=0.5\linewidth]{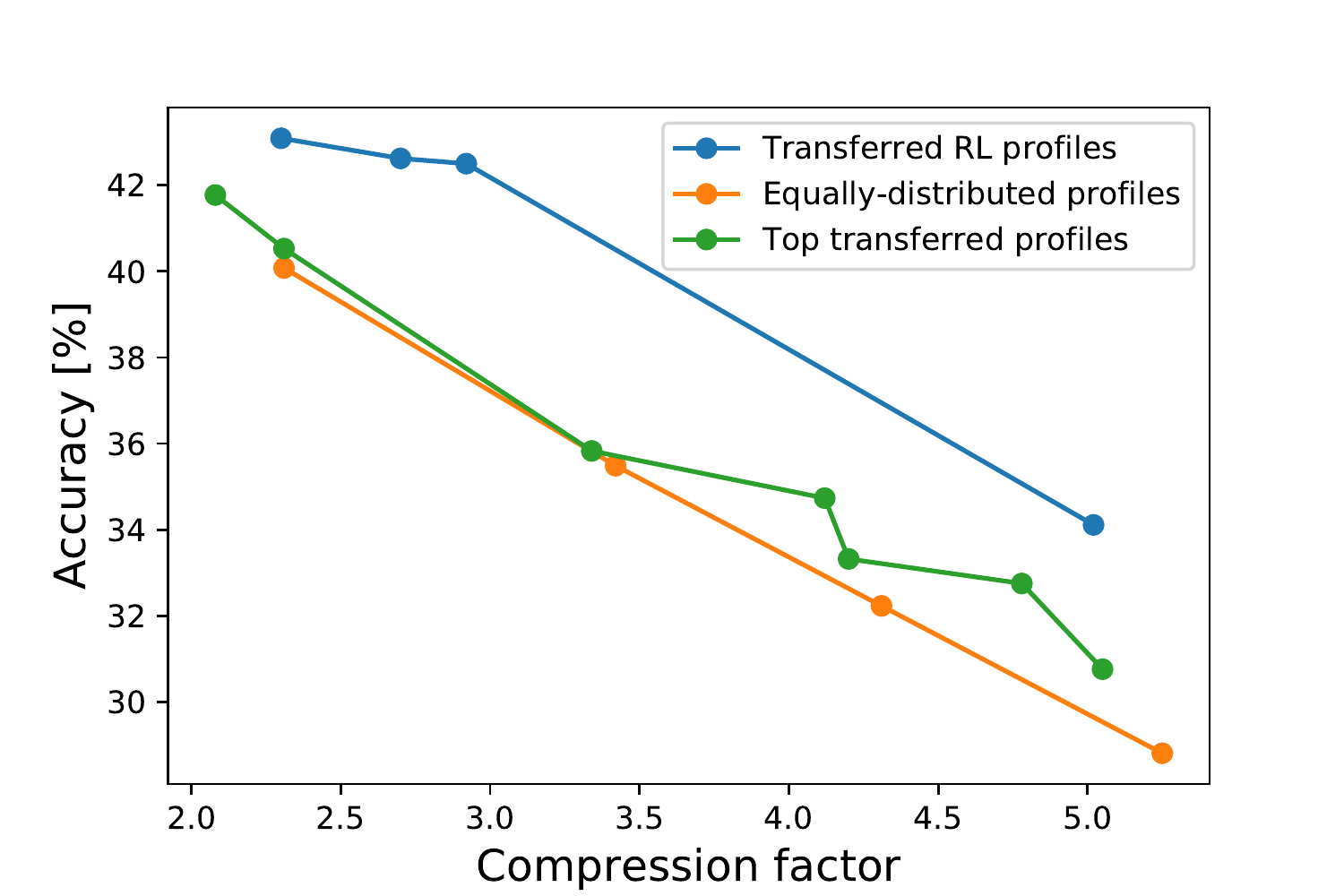}
	\includegraphics[width=0.5\linewidth]{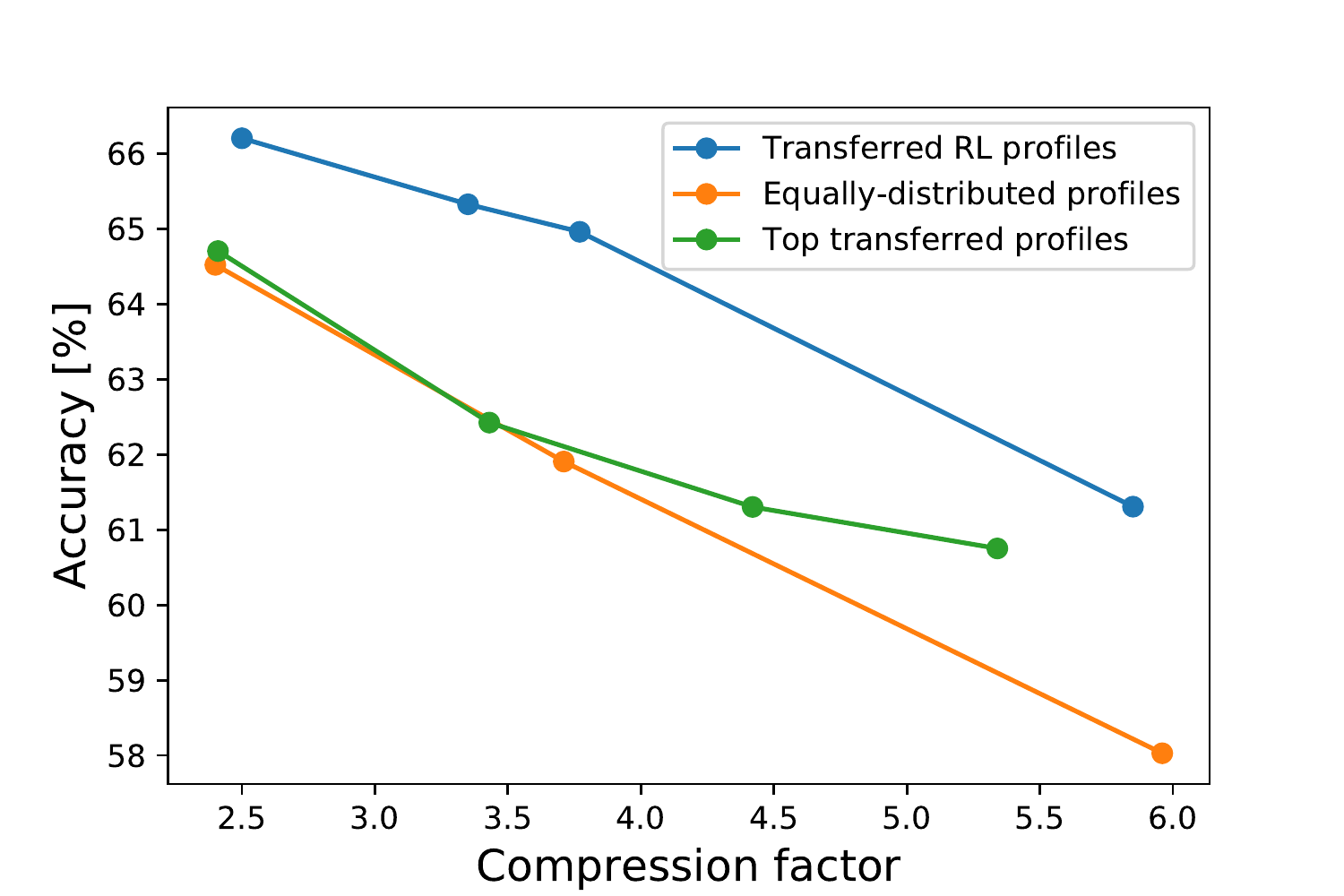}
	\caption{Out-of-the-box layer-wise profiles applied to Imagenet. Left: Resnet-20-16, Right: Resnet-20-64.}
	\label{fig:imagenet}	    
\end{figure*}

When transferring the same Cifar10 and Cifar100 profiles to a completely new  dataset - TinyImagenet, we discover that the profiles transferred from RL-search are significantly better than top profiles discovered from random-search.
Furthermore, we transfer the same profiles to Imagenet.
Not only is the Imagenet dataset different form Cifar10/Cifar100, the images are larger and the dataset is also much harder.
While we cannot produce massive amounts of randomly searched models for Imagenet, due to the cost-prohibitive nature of running hundreds of training jobs, we compare the RL-discovered profiles against the random profiles discovered to be the best from the exhaustive search and equally-distributed random profiles $P_e$.
The RL-based profile significantly out-performs the baselines at all CFs.

We also make a comparison against a state of the art iterative compression approach for channel-pruning: Differentiable Architecture Compression (DARC)~\cite{singh2019darc}.
DARC is a method that is fine-tuned for compression of one network on Imagenet.
While our RL-transferred Resnet-20 has a 5\% accuracy drop at 3.35x compression,  DARC, has a 3.7\% accuracy drop at 3.17x compression with the same network on Imagenet. 
This suggests that our out-of-the-box RL profiles can achieve results comparable with DARC which, consumes significantly more computation in identifying pruned profiles on target network.
This further demonstrates the strength of transferable profiles.

\noindent\textbf{Key insight from section \ref{sec:case}:} Using the proposed RL formulation to focus the search to,
\begin{enumerate*}
	\item operate around an expected compression ratio and, 
	\item maximize accuracy for multiple artifacts simultaneously,
\end{enumerate*}
provides us with transferable pruning profiles that generalize to datasets not used in the searching process.

\section{Conclusions}
\label{sec:conclusions}

In recent years, convolutional neural networks have ballooned in their sizes.
This is not scalable as larger compute resources are demanded as the popularity of training and inferencing CNNs also grow.
Channel pruning is therefore an important area of research.
In this work, we considered the most common baselines in channel pruning and deconstructed them taxonomically.
This lead us to isolate a few key stages and identify baseline methods for each stage.
We conducted various control studies at each stage and arrived at a few key insights:
\begin{enumerate*}
	\item We concluded that given fine-tuning, the most interesting search space for channel pruning is the profile of how much to prune per-layer.
	\item We concluded that pruning profiles are transferable across datasets.
\end{enumerate*}
Using these insights, we developed an RL-based search that can find profiles at a targeted compression factor that generalizes to unseen networks. 
We showed that pruning profiles found using such an RL-based search can transfer better than those found through exhaustive search.

\bibliographystyle{splncs04}
\bibliography{bib}

\begin{thebibliography}{10}
\providecommand{\url}[1]{\texttt{#1}}
\providecommand{\urlprefix}{URL }
\providecommand{\doi}[1]{https://doi.org/#1}

\bibitem{NIPS2016_6372}
Alvarez, J.M., Salzmann, M.: Learning the number of neurons in deep networks.
  In: Lee, D.D., Sugiyama, M., Luxburg, U.V., Guyon, I., Garnett, R. (eds.)
  Advances in Neural Information Processing Systems 29, pp. 2270--2278. Curran
  Associates, Inc. (2016),
  \url{http://papers.nips.cc/paper/6372-learning-the-number-of-neurons-in-deep-networks.pdf}

\bibitem{andrychowicz2020learning}
Andrychowicz, O.M., Baker, B., Chociej, M., Jozefowicz, R., McGrew, B.,
  Pachocki, J., Petron, A., Plappert, M., Powell, G., Ray, A., et~al.: Learning
  dexterous in-hand manipulation. The International Journal of Robotics
  Research  \textbf{39}(1),  3--20 (2020)

\bibitem{ashok2018nn}
Ashok, A., Rhinehart, N., Beainy, F., Kitani, K.M.: N2n learning: Network to
  network compression via policy gradient reinforcement learning. In:
  International Conference on Learning Representations (2018),
  \url{https://openreview.net/pdf?id=B1hcZZ-AW}

\bibitem{ashouri2018fast}
Ashouri, A.H., Abdelrahman, T.S., Remedios, A.D.: Fast on-the-fly
  retraining-free sparsification of convolutional neural networks. arXiv
  preprint arXiv:1811.04199  (2018)

\bibitem{berner2019dota}
Berner, C., Brockman, G., Chan, B., Cheung, V., Debiak, P., Dennison, C.,
  Farhi, D., Fischer, Q., Hashme, S., Hesse, C., et~al.: Dota 2 with large
  scale deep reinforcement learning. arXiv preprint arXiv:1912.06680  (2019)

\bibitem{cai2017reinforcement}
Cai, H., Chen, T., Zhang, W., Yu, Y., Wang, J.: Reinforcement learning for
  architecture search by network transformation. arXiv preprint
  arXiv:1707.04873  (2017)

\bibitem{cai2018proxylessnas}
Cai, H., Zhu, L., Han, S.: Proxylessnas: Direct neural architecture search on
  target task and hardware. arXiv preprint arXiv:1812.00332  (2018)

\bibitem{deng2009imagenet}
Deng, J., Dong, W., Socher, R., Li, L.J., Li, K., Fei-Fei, L.: Imagenet: A
  large-scale hierarchical image database  (2009)

\bibitem{frankle2018lottery}
Frankle, J., Carbin, M.: The lottery ticket hypothesis: Finding sparse,
  trainable neural networks. arXiv preprint arXiv:1803.03635  (2018)

\bibitem{gomez2019learning}
Gomez, A.N., Zhang, I., Swersky, K., Gal, Y., Hinton, G.E.: Learning sparse
  networks using targeted dropout. arXiv preprint arXiv:1905.13678  (2019)

\bibitem{NIPS2015_5784}
Han, S., Pool, J., Tran, J., Dally, W.: Learning both weights and connections
  for efficient neural network. In: Cortes, C., Lawrence, N.D., Lee, D.D.,
  Sugiyama, M., Garnett, R. (eds.) Advances in Neural Information Processing
  Systems 28, pp. 1135--1143. Curran Associates, Inc. (2015),
  \url{http://papers.nips.cc/paper/5784-learning-both-weights-and-connections-for-efficient-neural-network.pdf}

\bibitem{he2016deep}
He, K., Zhang, X., Ren, S., Sun, J.: Deep residual learning for image
  recognition. In: Proceedings of the IEEE conference on computer vision and
  pattern recognition. pp. 770--778 (2016)

\bibitem{he2018soft}
He, Y., Kang, G., Dong, X., Fu, Y., Yang, Y.: Soft filter pruning for
  accelerating deep convolutional neural networks. arXiv preprint
  arXiv:1808.06866  (2018)

\bibitem{he2018amc}
He, Y., Lin, J., Liu, Z., Wang, H., Li, L.J., Han, S.: Amc: Automl for model
  compression and acceleration on mobile devices. In: Proceedings of the
  European Conference on Computer Vision (ECCV). pp. 784--800 (2018)

\bibitem{he2017iccv}
He, Y., Zhang, X., Sun, J.: Channel pruning for accelerating very deep neural
  networks. In: The IEEE International Conference on Computer Vision (ICCV)
  (Oct 2017)

\bibitem{hu2016network}
Hu, H., Peng, R., Tai, Y.W., Tang, C.K.: Network trimming: A data-driven neuron
  pruning approach towards efficient deep architectures. arXiv preprint
  arXiv:1607.03250  (2016)

\bibitem{Huang_2018_ECCV}
Huang, Z., Wang, N.: Data-driven sparse structure selection for deep neural
  networks. In: The European Conference on Computer Vision (ECCV) (September
  2018)

\bibitem{hwangbo2019learning}
Hwangbo, J., Lee, J., Dosovitskiy, A., Bellicoso, D., Tsounis, V., Koltun, V.,
  Hutter, M.: Learning agile and dynamic motor skills for legged robots.
  Science Robotics  \textbf{4}(26),  eaau5872 (2019)

\bibitem{krizhevsky2010convolutional}
Krizhevsky, A., Hinton, G.: Convolutional deep belief networks on cifar-10.
  Unpublished manuscript  \textbf{40}(7) (2010)

\bibitem{krizhevsky2009learning}
Krizhevsky, A., Hinton, G., et~al.: Learning multiple layers of features from
  tiny images. Tech. rep., Citeseer (2009)

\bibitem{lebedev2016fast}
Lebedev, V., Lempitsky, V.: Fast convnets using group-wise brain damage. In:
  Proceedings of the IEEE Conference on Computer Vision and Pattern
  Recognition. pp. 2554--2564 (2016)

\bibitem{lecun1990optimal}
LeCun, Y., Denker, J.S., Solla, S.A.: Optimal brain damage. In: Advances in
  neural information processing systems. pp. 598--605 (1990)

\bibitem{tiny}
Li, F.F., Karpathy, A., Johnson, J.: Tiny imagenet. CS 231N: Standford
  University, http://cs231n.stanford.edu/  (2016)

\bibitem{rllib}
Liang, E., Liaw, R., Moritz, P., Nishihara, R., Fox, R., Goldberg, K.,
  Gonzalez, J.E., Jordan, M.I., Stoica, I.: Rllib: Abstractions for distributed
  reinforcement learning. arXiv preprint arXiv:1712.09381  (2017)

\bibitem{lillicrap2015continuous}
Lillicrap, T.P., Hunt, J.J., Pritzel, A., Heess, N., Erez, T., Tassa, Y.,
  Silver, D., Wierstra, D.: Continuous control with deep reinforcement
  learning. arXiv preprint arXiv:1509.02971  (2015)

\bibitem{lin2017nips}
Lin, J., Rao, Y., Lu, J., Zhou, J.: Runtime neural pruning. In: Advances in
  Neural Information Processing Systems. pp. 2181--2191 (2017),
  \url{http://papers.nips.cc/paper/6813-runtime-neural-pruning.pdf}

\bibitem{liu2017learning}
Liu, Z., Li, J., Shen, Z., Huang, G., Yan, S., Zhang, C.: Learning efficient
  convolutional networks through network slimming. In: Proceedings of the IEEE
  International Conference on Computer Vision. pp. 2736--2744 (2017)

\bibitem{liu2018rethinking}
Liu, Z., Sun, M., Zhou, T., Huang, G., Darrell, T.: Rethinking the value of
  network pruning. arXiv preprint arXiv:1810.05270  (2018)

\bibitem{louizos2017learning}
Louizos, C., Welling, M., Kingma, D.P.: Learning sparse neural networks through
  $ l\_0 $ regularization. arXiv preprint arXiv:1712.01312  (2017)

\bibitem{Luo_2017_ICCV}
Luo, J.H., Wu, J., Lin, W.: Thinet: A filter level pruning method for deep
  neural network compression. In: The IEEE International Conference on Computer
  Vision (ICCV) (Oct 2017)

\bibitem{mittal2018recovering}
Mittal, D., Bhardwaj, S., Khapra, M.M., Ravindran, B.: Recovering from random
  pruning: On the plasticity of deep convolutional neural networks. In: 2018
  IEEE Winter Conference on Applications of Computer Vision (WACV). pp.
  848--857. IEEE (2018)

\bibitem{molchanov2017variational}
Molchanov, D., Ashukha, A., Vetrov, D.: Variational dropout sparsifies deep
  neural networks. In: Proceedings of the 34th International Conference on
  Machine Learning-Volume 70. pp. 2498--2507. JMLR. org (2017)

\bibitem{molchanov2016pruning}
Molchanov, P., Tyree, S., Karras, T., Aila, T., Kautz, J.: Pruning
  convolutional neural networks for resource efficient inference. In:
  International Conference on Learning Representations (2016)

\bibitem{morcos2019one}
Morcos, A.S., Yu, H., Paganini, M., Tian, Y.: One ticket to win them all:
  generalizing lottery ticket initializations across datasets and optimizers.
  arXiv preprint arXiv:1906.02773  (2019)

\bibitem{pham2018efficient}
Pham, H., Guan, M.Y., Zoph, B., Le, Q.V., Dean, J.: Efficient neural
  architecture search via parameter sharing. arXiv preprint arXiv:1802.03268
  (2018)

\bibitem{qin2018demystifying}
Qin, Z., Yu, F., Liu, C., Chen, X.: Demystifying neural network filter pruning.
  arXiv preprint arXiv:1811.02639  (2018)

\bibitem{rajeswaran2017learning}
Rajeswaran, A., Kumar, V., Gupta, A., Vezzani, G., Schulman, J., Todorov, E.,
  Levine, S.: Learning complex dexterous manipulation with deep reinforcement
  learning and demonstrations. arXiv preprint arXiv:1709.10087  (2017)

\bibitem{sagemaker2019amazon}
SageMaker, A.: Amazon sagemaker (2019)

\bibitem{schulman2015high}
Schulman, J., Moritz, P., Levine, S., Jordan, M., Abbeel, P.: High-dimensional
  continuous control using generalized advantage estimation. arXiv preprint
  arXiv:1506.02438  (2015)

\bibitem{schulman2017proximal}
Schulman, J., Wolski, F., Dhariwal, P., Radford, A., Klimov, O.: Proximal
  policy optimization algorithms. arXiv preprint arXiv:1707.06347  (2017)

\bibitem{silver2016mastering}
Silver, D., Huang, A., Maddison, C.J., Guez, A., Sifre, L., Van Den~Driessche,
  G., Schrittwieser, J., Antonoglou, I., Panneershelvam, V., Lanctot, M.,
  et~al.: Mastering the game of go with deep neural networks and tree search.
  nature  \textbf{529}(7587), ~484 (2016)

\bibitem{singh2019darc}
Singh, S., Khetan, A., Karnin, Z.: Darc: Differentiable architecture
  compression. arXiv preprint arXiv:1905.08170  (2019)

\bibitem{srinivas2015data}
Srinivas, S., Babu, R.V.: Data-free parameter pruning for deep neural networks.
  arXiv preprint arXiv:1507.06149  (2015)

\bibitem{xavier}
Suau, X., Zappella, L., Palakkode, V., Apostoloff, N.: Principal filter
  analysis for guided network compression. CoRR  \textbf{abs/1807.10585}
  (2018), \url{http://arxiv.org/abs/1807.10585}

\bibitem{tian2017elf}
Tian, Y., Gong, Q., Shang, W., Wu, Y., Zitnick, C.L.: Elf: An extensive,
  lightweight and flexible research platform for real-time strategy games. In:
  Advances in Neural Information Processing Systems. pp. 2659--2669 (2017)

\bibitem{venkatesan2016diving}
Venkatesan, R., Li, B.: Diving deeper into mentee networks. arXiv preprint
  arXiv:1604.08220  (2016)

\bibitem{vinyals2019alphastar}
Vinyals, O., Babuschkin, I., Chung, J., Mathieu, M., Jaderberg, M., Czarnecki,
  W.M., Dudzik, A., Huang, A., Georgiev, P., Powell, R., et~al.: Alphastar:
  Mastering the real-time strategy game starcraft ii. DeepMind blog p.~2 (2019)

\bibitem{Wang_2018_ECCV}
Wang, X., Yu, F., Dou, Z.Y., Darrell, T., Gonzalez, J.E.: Skipnet: Learning
  dynamic routing in convolutional networks. In: The European Conference on
  Computer Vision (ECCV) (September 2018)

\bibitem{NIPS2016_6504}
Wen, W., Wu, C., Wang, Y., Chen, Y., Li, H.: Learning structured sparsity in
  deep neural networks. In: Advances in Neural Information Processing Systems
  29, pp. 2074--2082. Curran Associates, Inc. (2016),
  \url{http://papers.nips.cc/paper/6504-learning-structured-sparsity-in-deep-neural-networks.pdf}

\bibitem{ye2018rethinking}
Ye, J., Lu, X., Lin, Z., Wang, J.Z.: Rethinking the
  smaller-norm-less-informative assumption in channel pruning of convolution
  layers. arXiv preprint arXiv:1802.00124  (2018)

\bibitem{yu2019network}
Yu, J., Huang, T.: Network slimming by slimmable networks: Towards one-shot
  architecture search for channel numbers. arXiv preprint arXiv:1903.11728
  (2019)

\bibitem{yu2019universally}
Yu, J., Huang, T.: Universally slimmable networks and improved training
  techniques. arXiv preprint arXiv:1903.05134  (2019)

\bibitem{Yu_2018_CVPR}
Yu, R., Li, A., Chen, C.F., Lai, J.H., Morariu, V.I., Han, X., Gao, M., Lin,
  C.Y., Davis, L.S.: Nisp: Pruning networks using neuron importance score
  propagation. In: The IEEE Conference on Computer Vision and Pattern
  Recognition (CVPR) (June 2018)

\bibitem{zhou2016less}
Zhou, H., Alvarez, J.M., Porikli, F.: Less is more: Towards compact cnns. In:
  European Conference on Computer Vision. pp. 662--677. Springer (2016)

\bibitem{zhou2019deconstructing}
Zhou, H., Lan, J., Liu, R., Yosinski, J.: Deconstructing lottery tickets:
  Zeros, signs, and the supermask. arXiv preprint arXiv:1905.01067  (2019)

\bibitem{zhu2017prune}
Zhu, M., Gupta, S.: To prune, or not to prune: exploring the efficacy of
  pruning for model compression. arXiv preprint arXiv:1710.01878  (2017)

\bibitem{zhuang2018discrimination}
Zhuang, Z., Tan, M., Zhuang, B., Liu, J., Guo, Y., Wu, Q., Huang, J., Zhu, J.:
  Discrimination-aware channel pruning for deep neural networks. In: Advances
  in Neural Information Processing Systems. pp. 883--894 (2018)

\bibitem{zoph2016neural}
Zoph, B., Le, Q.V.: Neural architecture search with reinforcement learning.
  arXiv preprint arXiv:1611.01578  (2016)

\end{thebibliography}

\section*{Supplementary}
\begin{figure}
	\centering
	\includegraphics[width=\linewidth]{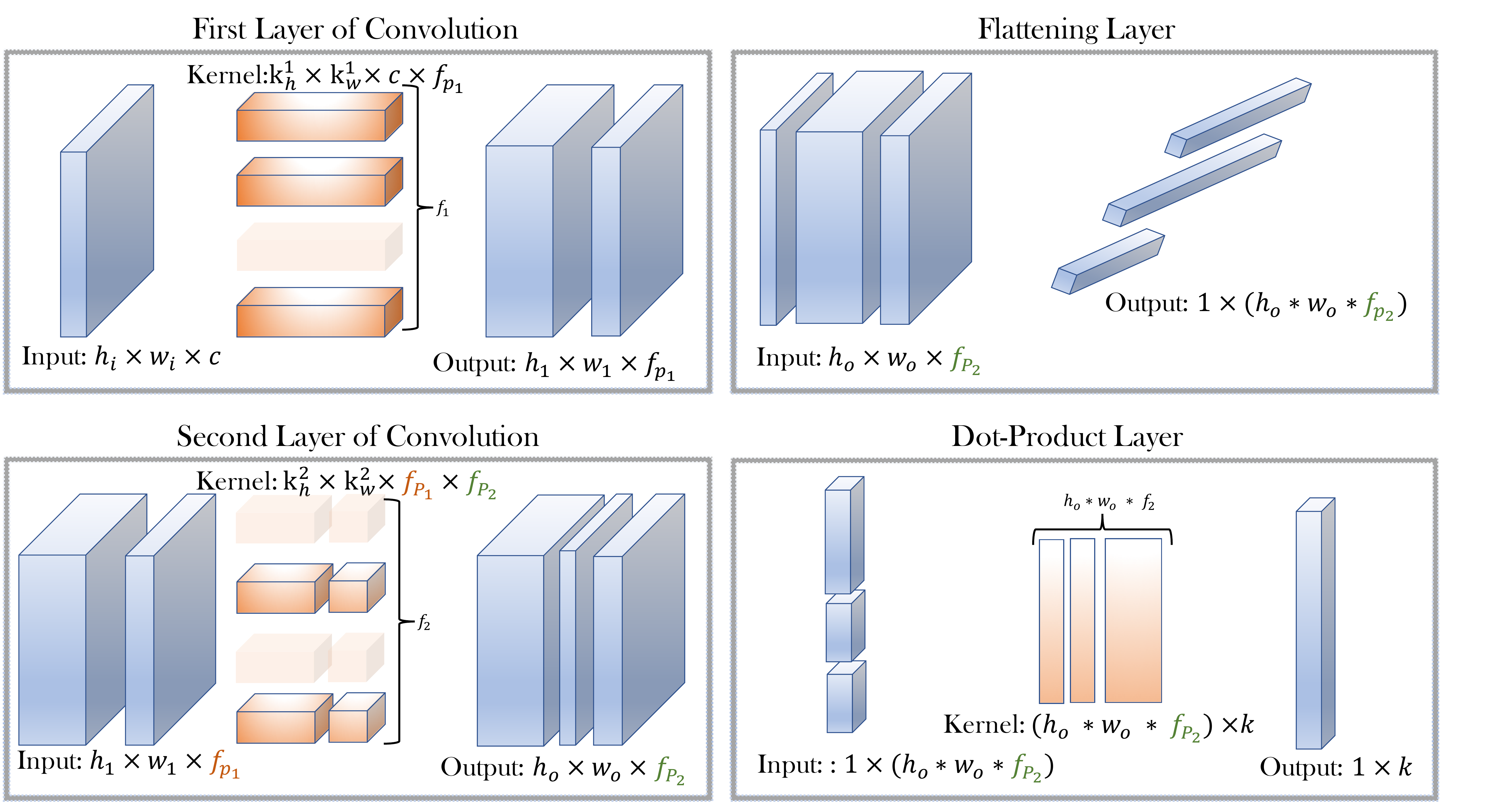}		
	\caption{An illustration of Channel Pruning.}	
	\label{fig:compression}	     	
\end{figure}

Consider an exemplar convolutional neural network $N$.
Suppose the network consume images $x \in \region^{h_i \times w_i \times c}$.
Suppose it consists of two convolutional layers with kernels $w_1 \in \region^{k_h^1 \times k_w^1 \times c \times f_1}$ and $w_2 \in \region^{k_h^2 \times k_w^2 \times f_1 \times f_2}$ producing output tensors $L_1 \in \region^{h_1 \times w_1 \times f_1}$ and $L_2 \in \region^{h_o \times w_o \times f_2}$ respectively.
Suppose that $L_2$ is flattened and supplied to a fully-connected layer $L_{\text{fc}}$ with weights $w_{\text{fc}} \in \region^{h_o*w_o*f_2 \times k}$ producing an output of $L_{\text{fc}} \in \region^{1 \times k}$ to classify the inputs into $k$ categories.
Consider the illustration of channel pruning in figure \ref{fig:compression}.
Retaining only $f_{p_1} \leq f_1$ channels in layer $L_1$ implies that in the layer $L_2$, for every channel, we need only $f_{p_1}$ filters.
The filters that convolve with the pruned channels of $L_1$ are no longer needed and can be removed.
Consequently, channel pruning on one layer performs filter pruning on the subsequent layer.		
Further pruning on the second layer leaves us with only $f_{p_2}$ channels, reducing the computational load on the fully-connected layer as well, which is now a dot-product with a tensor $w_{\text{fc}} \in \region^{h*w*f_{p_2} \times k}, f_{p_2} \leq f_2$.
Contingent on the ordering of flattening, columns of weights (filters) in $w_{\text{fc}}$ that multiply across the pruned channels in $L_2$ can also be removed.

Note that when we prune a network, we do not simply multiply zeros to the parameters and keep them frozen.
Instead we remove the dimensions of the tensor with zero-mask value.
We then reconstruct the tensor (and therefore the entire network) in memory with the new tensor.
Since the pruning for all layers are structured, the reconstructed tensors are not sparse, therefore a pruned convolutional layer is a typical full-convolutional layer.
This implies that we do not require any special framework-specific low-level implementations. 

\section{Architecture of networks used}

In this work, we have used three network architecture: C-NET, ResNet-20-16 and ResNet-20-64.
In this section, we will discuss these networks and any special pruning implementations that are of concern.

\subsection{Cylinder network}
C-NET or cylinder network is a network where we controlled the number of channels in each layer to be the same. 
This allowed us to study channel pruning without the effect of layer-specific properties such as bottleneck or batchnorms.
The C-NET architecture is illustrated in figure~\ref{fig:networks}.
Each layer contains $32$ channels. 
This implies that $\beta_i \in \bool^{32}$.
After every second convolutional layer, we maxpool by $2$.
All convolutional layers were also followed by a Relu activation layer.
The output of the last Relu layer was flattened before being used as input for dense layer.
The flattening process is performed using row-major index order\footnote{Also referred to as C-like order.}, where we unroll the last axis index first and so on.
This implies that we simply need to replicate $\alpha_5$ enough number of times to meet the incoming dimensions of the dense layer, to get $\bar{\alpha_5}$.
This process is deterministic and no new pruning decisions are made for the dense layer.
This network architecture requires 6 decisions.

\subsection{Residual Networks}
The ResNets we used are defined in figure~\ref{fig:networks}.
The only difference between the two networks are the number of channels, with ResNet-20-16 having blocks with channels $16, 32$ and $128$ and ResNet 20-64 having blocks with channels $64, 128$ and $512$ respectively.
Pruning ResNets are slightly more trickier than purning a simple feed-forward network.
ResNets present us with two major architectural problems:
\begin{enumerate}[wide, labelwidth=!, labelindent=0pt, nosep,before=\leavevmode]
	\item How do we  prune the Batchnorm layer?
	\item How do we prune the shortcuts?
\end{enumerate}

\subsubsection{Pruning batchnorm layers}
\begin{figure}
	\centering
	\includegraphics[height=0.9\textheight]{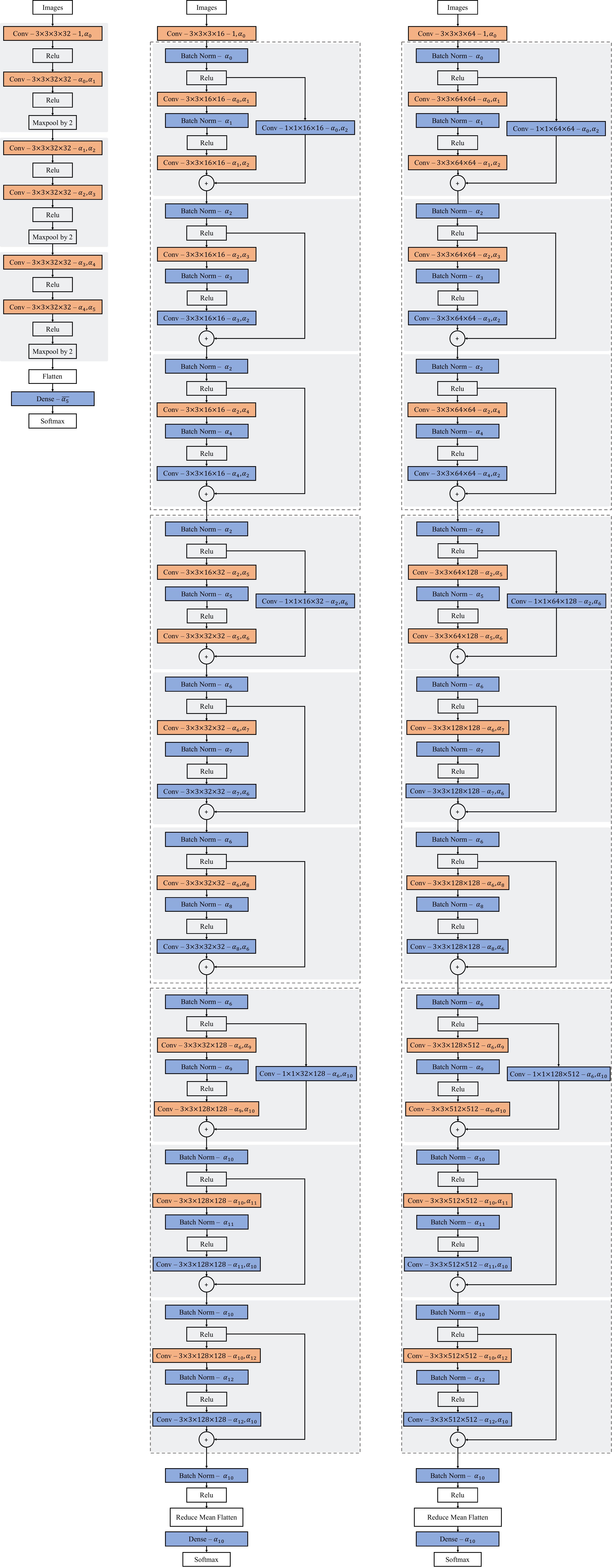}
	\caption{Illustration of the networks: C-NET, ResNet-20-16, ResNet-20-64 from left to right. Layers that are in orange are layers for which a unique pruning mask $\alpha$ is generated. Layers in blue simply get applied the same pruning mask from another ``orange" layer. For convolutional layers, we have noted the masks for dimension 3 and 4. For other layers, the mask is always the last dimension.}
	\label{fig:networks}
\end{figure}

Batchnorm layers are tied to the convolutional layers on the channel dimension.
This implies that a convolutional layer with a batch norm layer following it, can simply get the same pruning flag as the convolutional layer itself.
This implies that any batchnorm statistic learned across a particular channel dimension will remain and the others will simply be ignored.
Figure~\ref{fig:networks} illustrates how the prune flags are distributed to the batch norm layers.

\subsubsection{Projection shortcut blocks}
The first building block of any residual block in the ResNet architecture typically has a projection shortcut convolution layer.
These are needed because the first building block in every block typically increases the number of the channels, while the other building blocks maintain the number of channels.
These projection layers are $1 \times 1$ convolutional layers.
The purpose of these layers is to re-project the values in the residual path so that the dimensionality matches with the block's outgoing signal, which has more dimensions than the incoming signal.
We do this to ensure that the appropriate dimensions are also pruned to maintain element-wise consistency\footnote{Addition of the residue is performed element-wise, therefore dimensions must match.}.

Consider the first building block in the first block of the ResNets of figure~\ref{fig:networks}.
This block is one that has a projection shortcut.
We apply to the projection shortcuts third dimension, the same mask that enters the block from the previous layers $\alpha_0$.
The mask for the fourth dimension has to be dimensionally-locked with the outgoing mask of the building block.
For the case of the first building block of ResNets, it is $\alpha_2$.
Therefore, we apply $\alpha_2$ to the fourth dimension of the projection shortcut.
Note that projection shortcut is not actively pruned.
The pruning decisions for both dimension of the projection shortcut are determined by the other convolutional layers in the building block.

\subsubsection{Non-projection shortcut blocks}
The other type of building blocks in the ResNet are non-projection shortcut blocks.
Consider the second building block (in the first block) of the ResNets.
This block contains a shortcut, but there is no projection on this shortcut.
Considering that this is a simple residue addition, we can reconsider the dimensions that are retained in this shortcut as compared to the original path, as selected using $\alpha_2$.
Therefore at the convolutional layer at the end of the block, to which the residue gets added, we use the same pruning mask $\alpha_2$ for the outgoing dimension.
Therefore, the pruning decisions for all outgoing convolutional layers in non-projection shortcut building blocks were determined at the time of making the pruning decision for the outgoing convolutional layer of the immediately preceding building block with a projection shortcut.
Figure~\ref{fig:networks} illustrates the filter and channel masks for all convolutional layers and channel masks for all batch norm layers that were used for the entire network.

Putting all this together, gives us 13 actively prunable layers, for which our policy needs to make decisions, on the ResNet-20 architecture.
Table~\ref{tab:flags} shows the lengths of all the pruning flags used.
\begin{center}
	\begin{table}[!t]
		\centering
		\begin{tabular}{|c||c|c|c|}
			\hline
			\textbf{Flag}          & \textbf{C-NET} & \textbf{ResNet-20-16} & \textbf{ResNet-20-64} \\
			\hline\hline
			$\alpha_0$    & 32    & 16           & 64          \\
			$\alpha_1$    & 32    & 16           & 64          \\
			$\alpha_2$    & 32    & 16           & 64          \\
			$\alpha_3$    & 32    & 16           & 64          \\
			$\alpha_4$    & 32    & 16           & 64          \\
			$\alpha_5$    & 32    & 32           & 128         \\
			$\alpha_6$    &          & 32           & 128         \\
			$\alpha_7$    &          & 32           & 128         \\
			$\alpha_8$    &          & 32           & 128         \\
			$\alpha_9$    &         & 128          & 512         \\
			$\alpha_{10}$ &       & 128          & 512         \\
			$\alpha_{11}$ &       & 128          & 512         \\
			$\alpha_{12}$ &       & 128          & 512        \\
			\hline
		\end{tabular}
		\caption{All pruning flags used and their lengths.}
		\label{tab:flags}
	\end{table}
	\begin{table}[!t]
		\centering
		\begin{tabular}{|l|l||l|}
			\hline
			\textbf{Dataset}        & \textbf{Model} & \textbf{Accuracy} \\ \hline \hline
			\textit{Cifar10}  & \multirow{2}{*}{\textit{C-NET}}   			  &  82.39            \\ \cline{1-1}\cline{3-3}
			\textit{Cifar100} &          &  46.95           \\ \hline
			\textit{Cifar10}  & \multirow{4}{*}{\textit{ResNet-20-16}}  & 92.25             \\ \cline{1-1}\cline{3-3}
			\textit{Cifar100} &          & 68.85             \\ \cline{1-1}\cline{3-3}
			\textit{Tiny Imagenet} &          & 48.66 \\ \cline{1-1}\cline{3-3}
			\textit{Imagenet}  &          & 50.17 \\ \hline
			\textit{Cifar10}  & \multirow{3}{*}{\textit{ResNet-20-64}} & 94.38             \\ \cline{1-1}\cline{3-3}
			\textit{Cifar100} &          & 75.11             \\ \cline{1-1}\cline{3-3}
			\textit{Imagenet}&          & 68.78         \\ \hline
		\end{tabular}
		\caption{Base models and accuracies.}
		\label{tab:base}
	\end{table}
\end{center}

\section{Base models}

For all experiments we use a once-trained set of base-models.
Table~\ref{tab:base} provides all the base model's accuracies that we used.

\section{Reinforcement learning details}
\subsection{Observations}
To derive our features we follow an idea mirroring the ones proposed in~\cite{molchanov2016pruning}, and repurpose the channel importance metric therein.
Refer to the original paper ~\cite{molchanov2016pruning} for more details.
We essentially re-derive the same for our purpose here, for the sake of context and completeness.
Consider the output of the layer $L_t \in \region^{h \times w \times c_t}.$
Let, $L_t^i \in \region^{h \times w}$ be the $i^{\text{th}}$ channel of output implying that $L_t = \{L_t^1 \dots L_t^{c_t} \}$.
Analogously, $w_t = \{w_t^1 \dots w_t^{c_t} \}$ split across the output channels dimension.
Consider also, some overall objective function of the network itself $\epsilon(x)$.
Under assumptions of channel independence $\epsilon(x \vert L_t^i) = \epsilon(x \vert w_t^i)$.
Further assuming independence of filters within the channel, we can get the change in objective with a particular channel $i$ of the layer $t$ removed with $\alpha_i = 0$ as,

\begin{equation}
\vert \Delta (\epsilon (x \vert L_t^i)) \vert = \vert \epsilon(x \vert L_t^i = 0) - \epsilon(x, L_t^i) \vert.
\end{equation}
Approximating the objective $\epsilon(x \vert L_t^i)$ with a Taylor series first-order polynomial near $L_t^i = 0$ for a pruned channel $i$, we have:
\begin{equation}
\epsilon(x \vert L_t^i = 0) = \epsilon(x \vert L_t^i) - \frac{\partial \epsilon}{\partial
	L_t^i}L_t^i + \frac{\partial^2 \epsilon}{\partial^2 L_t^i} \frac{(L_t^i)^2}{2}.
\label{eqn:intermediate}
\end{equation}
The second-order term, $\frac{\partial^2 \epsilon}{\partial^2 L_t^i} \frac{(L_t^i)^2}{2}$ is basically the saliency term introduced by LeCun \etal and is a common term in itself used for filter pruning~\cite{lecun1990optimal}.
Substituting this in the equation \ref{eqn:intermediate}, we get,
\begin{align*}
\vert \Delta (\epsilon ({L_t^i})) \vert &= \Bigg\vert \epsilon(x,
L_t^i) -  \frac{\partial \epsilon}{\partial L_t^i}L_t^i + \frac{\partial^2
	\epsilon}{\partial^2 L_t^i} \frac{(L_t^i)^2}{2} - \epsilon(x, L_t^i)\Bigg\vert. \\
&= \Bigg\vert \frac{\partial^2\epsilon} {\partial^2
	L_t^i} \frac{(L_t^i)^2}{2} - \frac{\partial \epsilon}{\partial L_t^i}L_t^i\Bigg\vert.
\end{align*}
We ignore the Hessian term as it is too small and instead use $\Phi(L_t^i) =  \Big\vert\frac{\partial \epsilon} {\partial L_t^i} L_t^i \Big\vert$ as observations.
They are therefore an approximation to the consequence of removing a channel.

The observation is, for layer $L_t$ with $c_t$ channels, and a pruning action $\beta_t \in [0,1]$ the observations corresponding to channel $j$ are,
\begin{equation}
\Phi(L_t^j \vert \beta_t) =  \begin{cases} 
\expectation_{k_h, k_w, c_{t-1}} \Big \vert  \frac{\partial \epsilon} {\partial L_t^j} L_t^j \Big\vert  & \text{if } \Bernoulli(\beta_t)= 1 \\
0 & \text{otherwise}
\end{cases}.
\label{eqn:observations}
\end{equation}
This term is a direct consequence of change in the loss with respect to the actions that were just taken.

\subsection{Actions}

The action space for layer $L_t$ from the RL perspective is $\beta_t \in \region$, which the policy will decode using Bernoulli to a Boolean space $\bool^{c_t}$.
On the consumption of the observations $\Phi(.)$, the policy produces a Boolean set of actions $\alpha_t \in \bool^{c_t}$.
The consequence of these actions are that the channels are pruned in $w_t$ and appropriate filters are also back-pruned in $w_{t+1}$.
Implementationally, this could lead to singularities when $\vert\vert \alpha_t \vert\vert_1$ is very small.
To avoid this problem we pad the lower end of the action space 10\% forcing $\vert\vert \alpha_t \vert\vert_1> 0.1$.

\subsection{Rewards}
\begin{figure}[t]
	\includegraphics[width=0.325\linewidth]{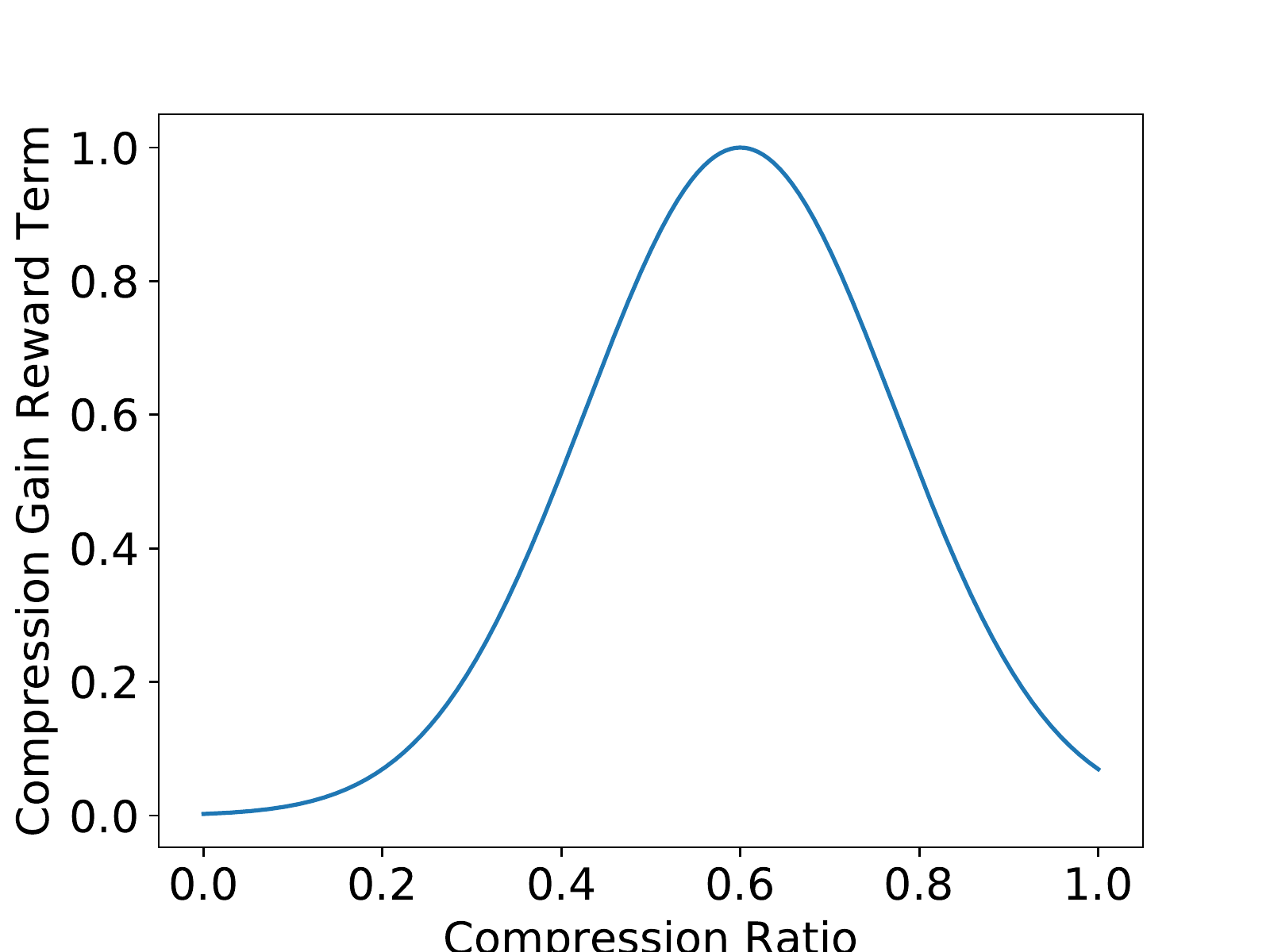}
	\includegraphics[width=0.325\linewidth]{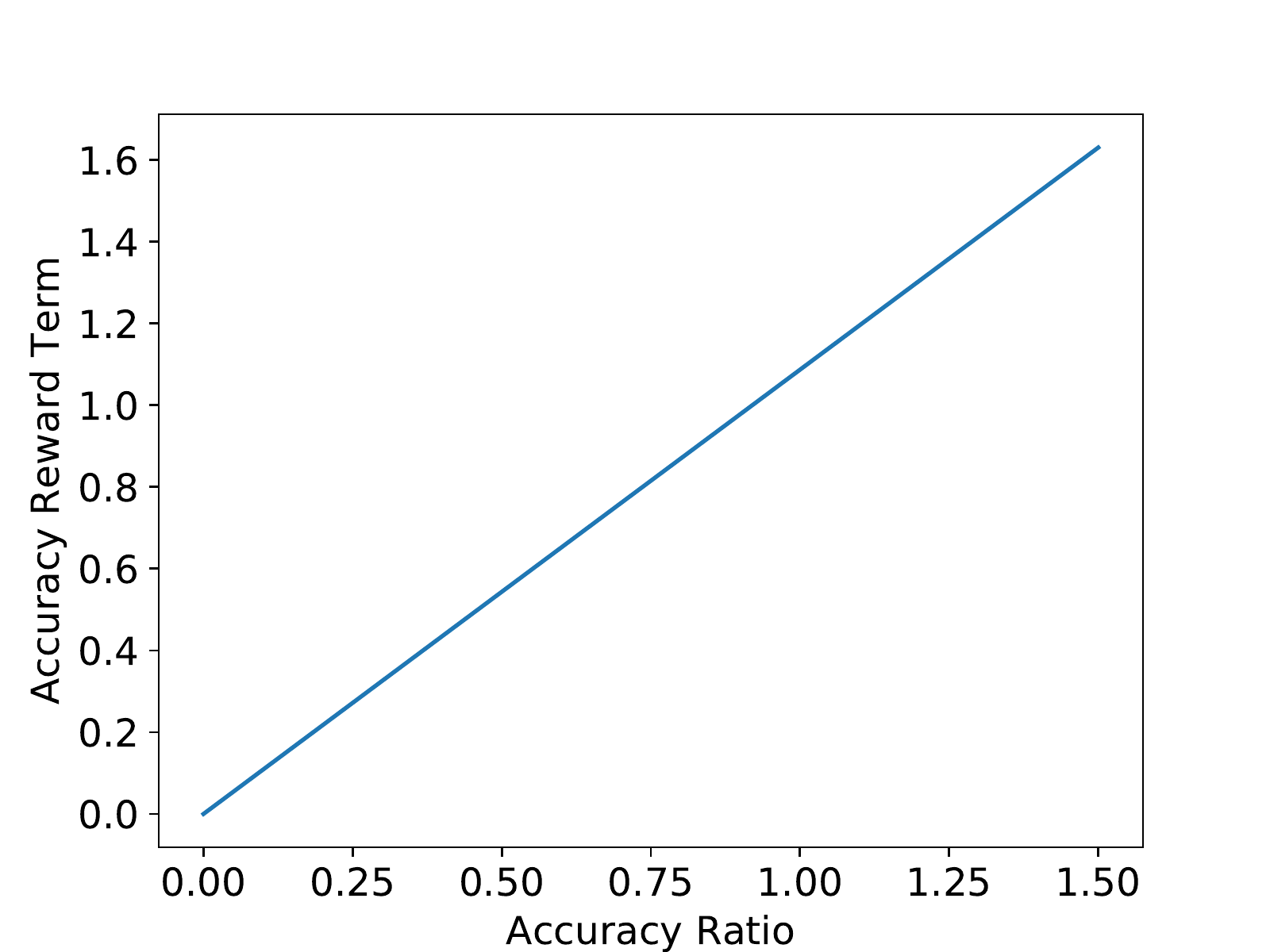}
	\includegraphics[width=0.325\linewidth]{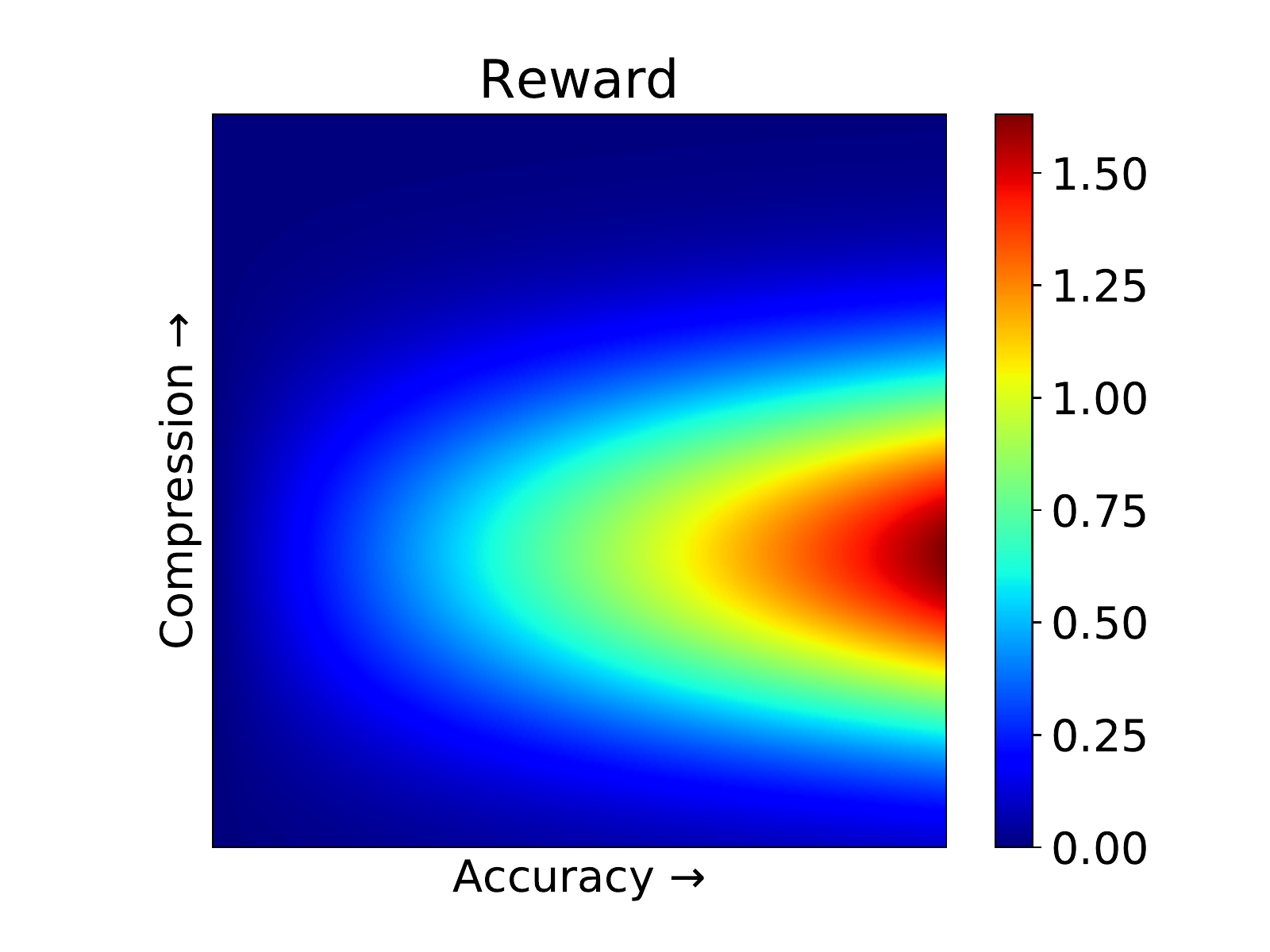} 		 
	\caption{An illustration of the proposed reward landscape. The expected compression is $C_e = 0.5$ and the original model's accuracy is $A_e=0.9$,  $\sigma=0.3$.}
	\label{fig:reward_landscape_7}	    
\end{figure}
The reward we used (equation 10 in main paper) for the end of episode case is is illustrated in figure~\ref{fig:reward_landscape_7}.
We refer to this as the Gaussian reward or expected compression reward.
This reward is optimal when the agent finds profiles closer to $C_e$.
So far as we are aware, we are first to use such a reward function. 
Consider for instance, the reward function used by Ashok \etal\, which is the closest analogy we have~\cite{ashok2018nn}.
\begin{equation}
r_t =  \begin{cases}
0 & \text{if} \ t \neq l \\
\frac{A}{A_e} (1-C)^2& \text{otherwise},
\end{cases}.
\label{eqn:reward}
\end{equation}
\begin{figure}[t]	
	\includegraphics[width=0.325\linewidth]{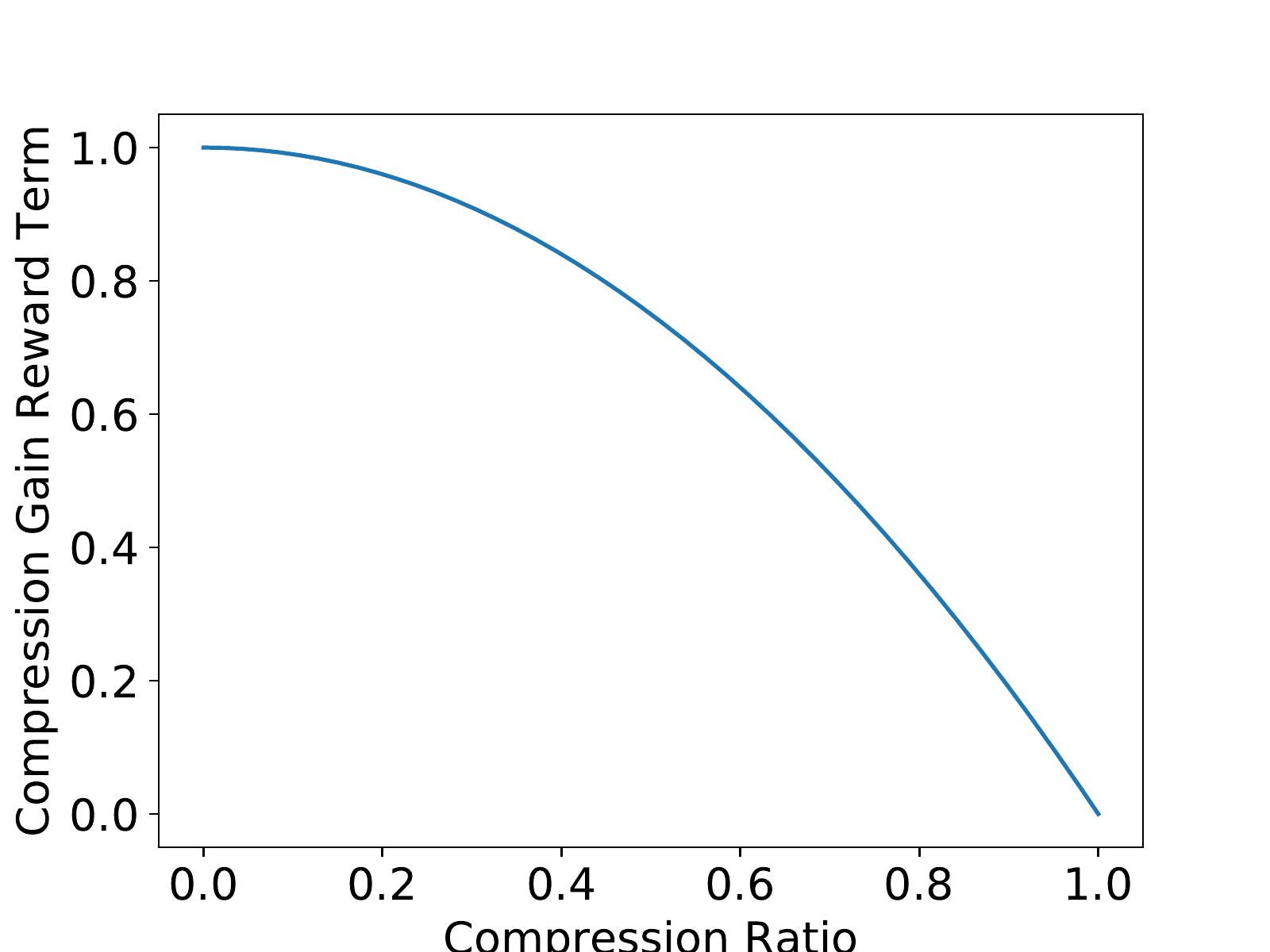}
	\includegraphics[width=0.325\linewidth]{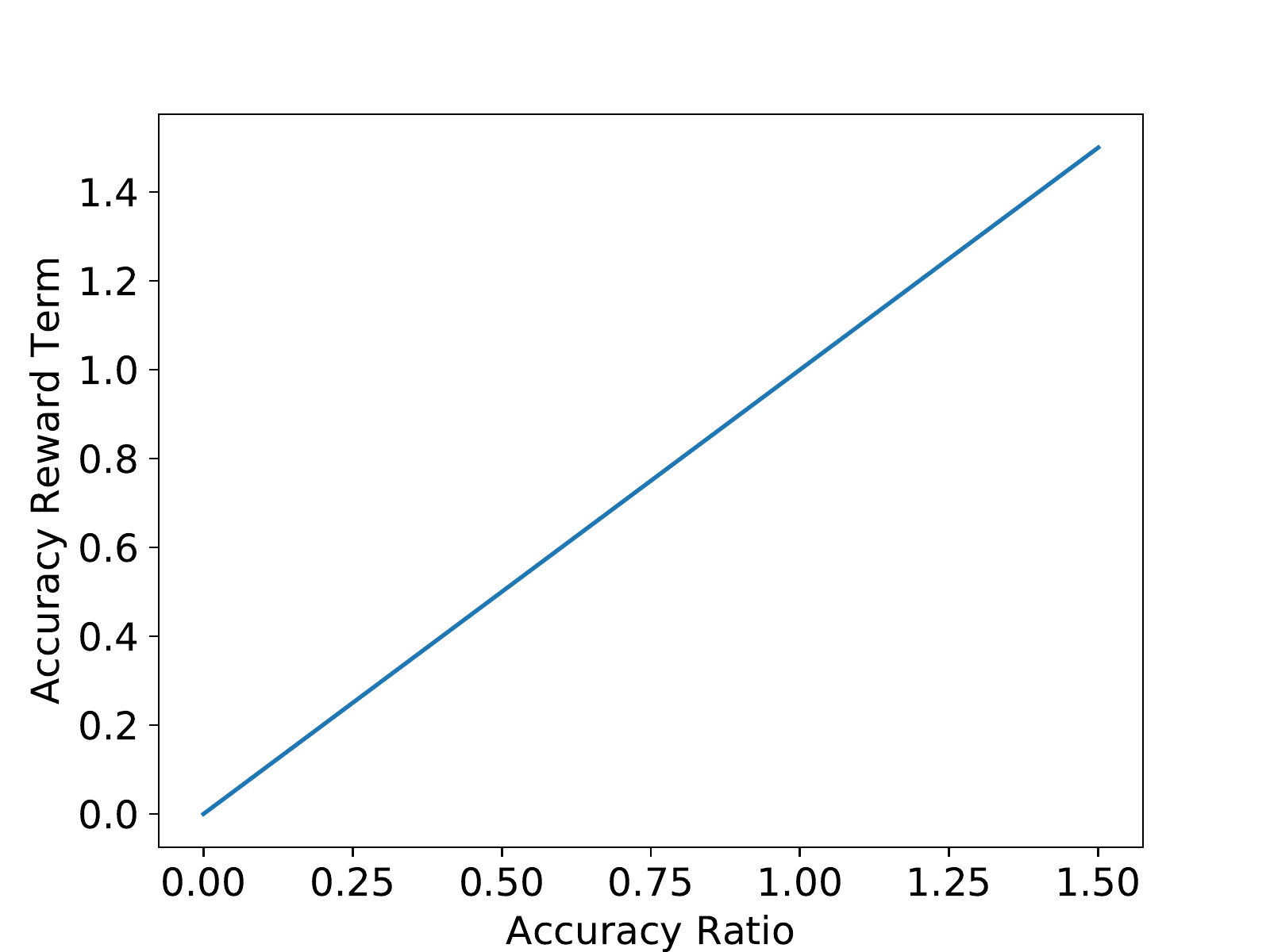}
	\includegraphics[width=0.325\linewidth]{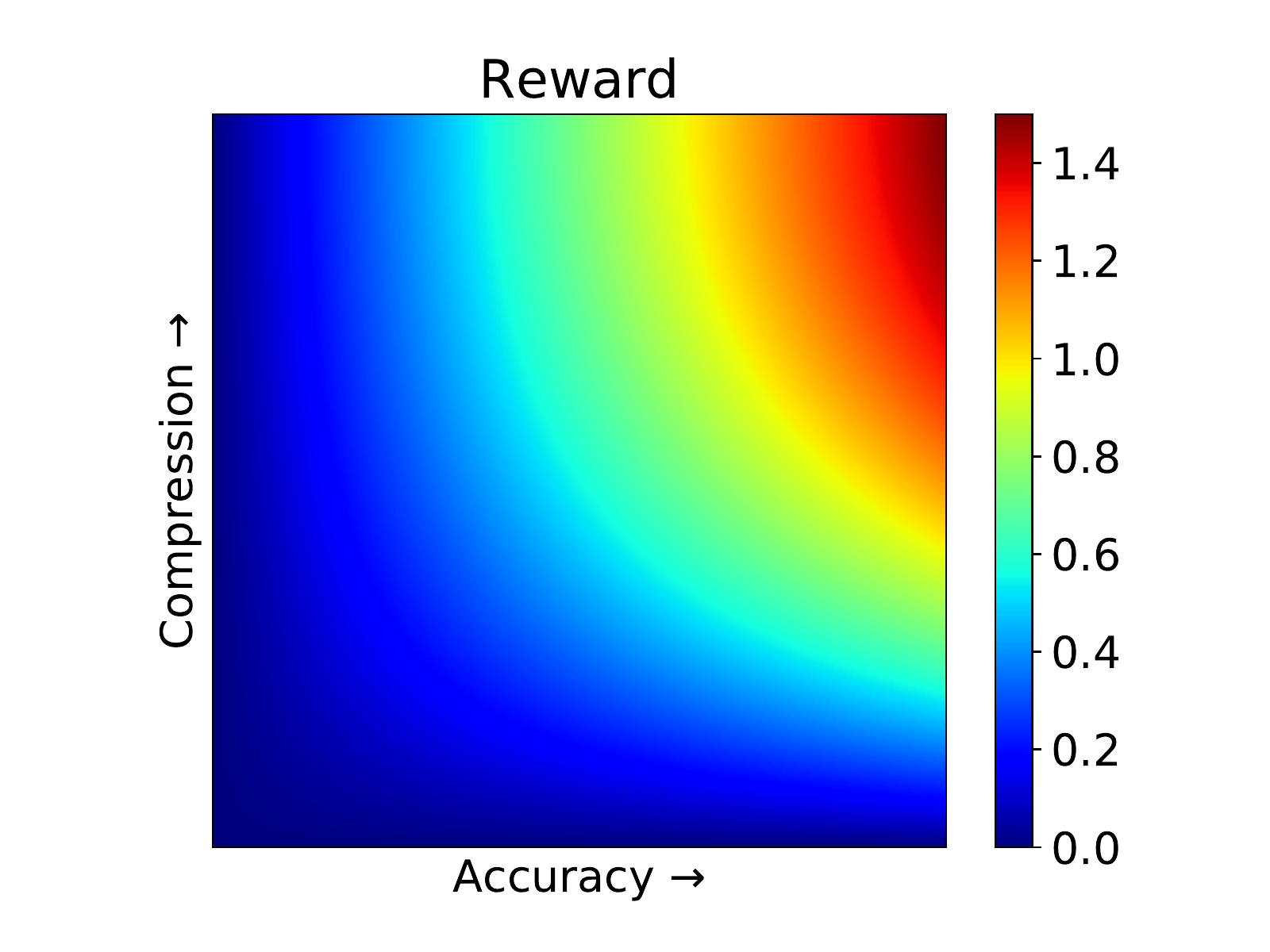} 		 
	\caption{An illustration of the reward landscape proposed by Ashok \etal\. The original model's accuracy is $A_e=0.9$.}
	\label{fig:reward_n2n}	    
\end{figure}
This reward is illustrated in figure~\ref{fig:reward_n2n}.
This reward leads to very inconsistent learning of the RL.
The agent under such a monotonic reward setting has the following two problems:
\begin{enumerate}
	\item The exploration space is completely unconstrained. The RL agent  doesn't get any guidance from the reward on which direction to explore, therefore there are no guarantees that it will even find an optimal compression factor. Even if the agent did, it takes a prohibitively large amount of time. This is not feasible when we want to train entire models from scratch inside our environments.
	
	\item We observe that when the actions are tightly related to the reward in a monotonic way, the agents simply tend to collapse into extreme or zero compression and never recovers from such a state.
	Essentially, the actions travel along the locus of constant reward in figure~\ref{fig:reward_n2n}.
\end{enumerate}

We also considered another related monotonic reward, which we call the hyperbolic reward, as follows:
The compression reward for pruning layer $L_t$ is,
\begin{equation}
r_c^t = \frac{\tanh \Big( \frac{1 - \frac{\sum \alpha_t}{c_t} - C_e}  \tau \Big) + \tanh\frac {C_e}{\tau}}
{\tanh \Big( \frac{1 -C_e}{\tau} \Big) + \tanh\frac {C_e}{\tau}},
\end{equation}
where, $\alpha_t$ is the prune policy for that layer and $C_e$ is a tolerance below which we want to discourage our agent dramatically and $\tau$ is a constant temperature value to stretch the hyperbole.
The accuracy reward for pruning layer $L_t$ is,
\begin{equation}
r_a^t = \frac{\tanh \Big( \frac{\frac{\sum a_t}{A} - A_e}  \tau \Big) + \tanh\frac {A_e}{\tau}}
{\tanh \Big( \frac{1 -A_e}{\tau} \Big) + \tanh\frac {A_e}{\tau}},
\end{equation}
where, $a_t$ is the accuracy on the validation set after pruning the $t^{\text{th}}$ layer, $A$ is an accuracy normalizer, typically the accuracy of the unpruned base network and $A_e$ is the expected accuracy.
The final reward is,
\begin{equation}
r_t =  \begin{cases}
0 & \text{if} \ t \neq l \\
r_a^l\sum_{i=1}^{l} r_c^l & \text{otherwise},
\end{cases}.
\label{eqn:reward_hp}
\end{equation}

\begin{figure}	
	\includegraphics[width=0.325\linewidth]{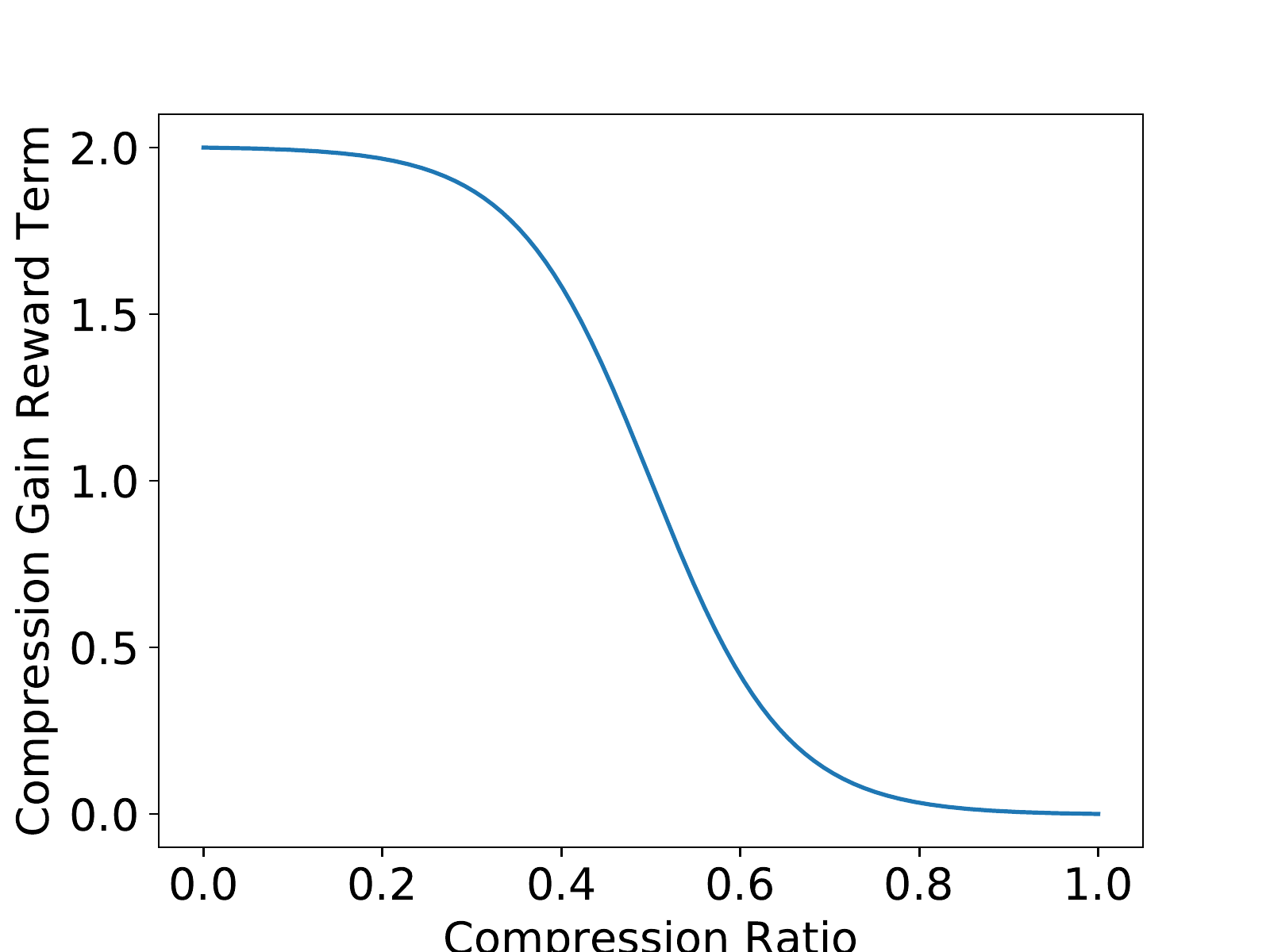}
	\includegraphics[width=0.325\linewidth]{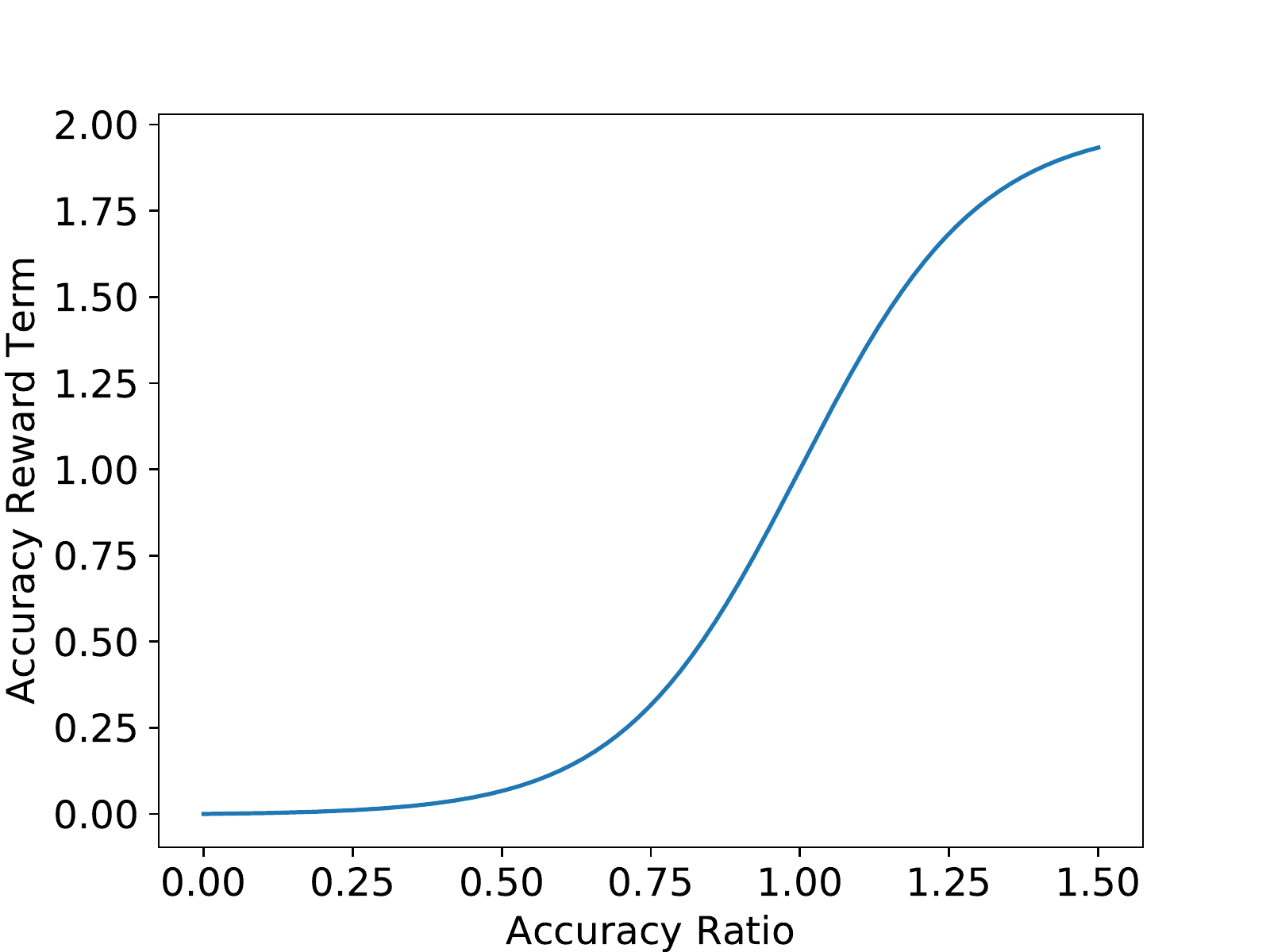}
	\includegraphics[width=0.325\linewidth]{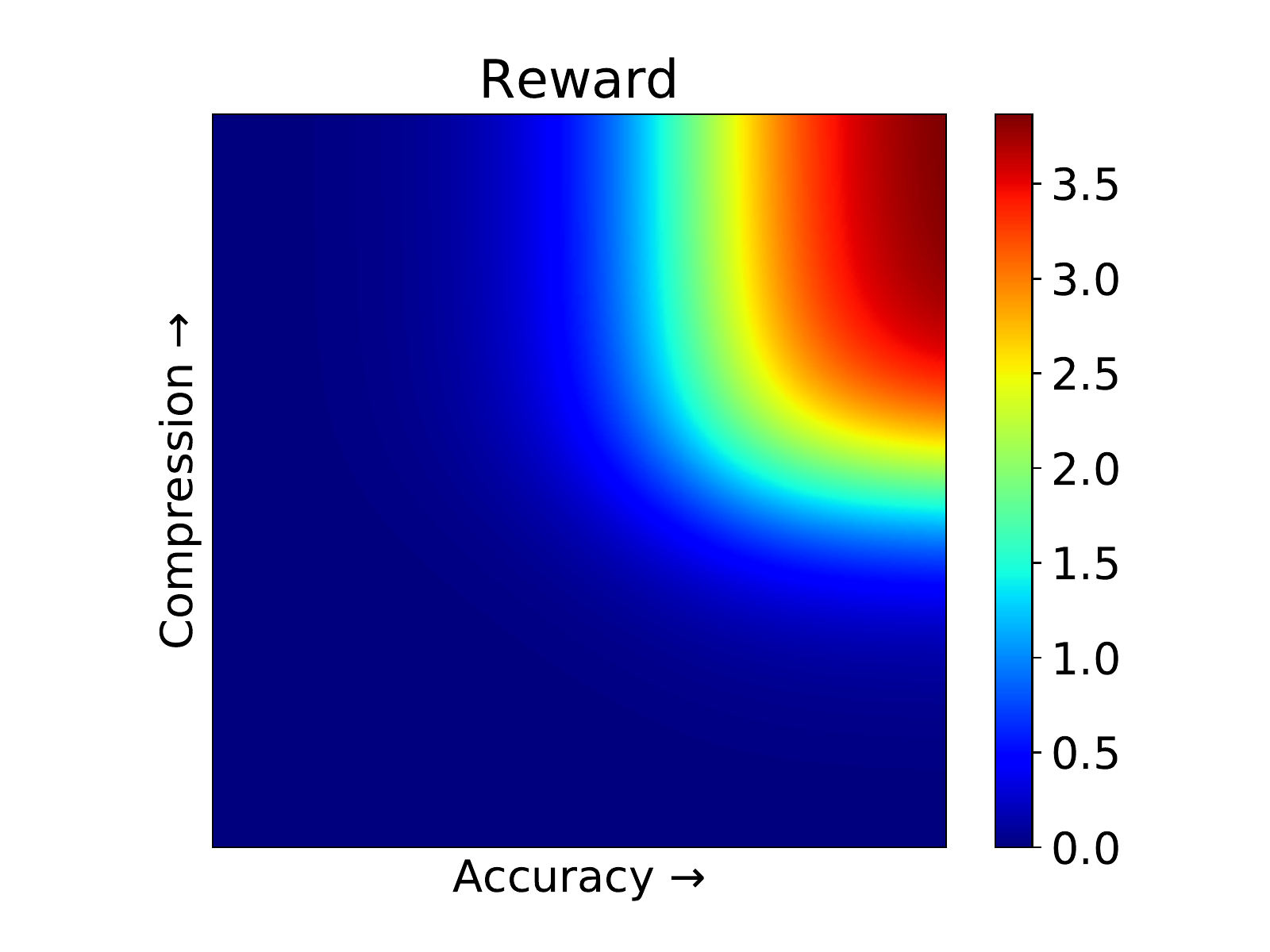} 		 
	\caption{An illustration of the hyperbolic reward landscape. $A_e=0.9, C_e=0.8$.}
	\label{fig:reward_hyperbolic}	    
\end{figure}
Figure \ref{fig:reward_hyperbolic} illustrates the hyperbolic reward landscape. 
It can be seen how the hyperbolic rewards encourage the agents to maintain accuracy and compression above our expectation as a primary goal and then force them both higher.
Despite this, similar problems exist and we find that the Gaussian reward performs the best throughout the experiments.

\subsection{Distributed  RL training}
The major challenge of training a RL policy comes from the fact that the simulation environment is constructed based on a complex neural network. 
At each time step, the network needs to be re-trained after the agent takes an action. 
Hence collecting enough simulation episodes is computationally expensive for the agent to learn a reasonale policy.
To adress this, we leverage Ray RLlib \cite{rllib} and Amazon SageMaker \cite{sagemaker2019amazon}.
Specifically, we use the API provided by RLlib which implements parallel simulation and optimization of such RL problem.
To enable efficient resource allocation and utilization, we utilize Amazon SageMaker to start a Ray cluster and perform distributed training using on demand instances.
In this section, we provide some psuedo-code for further simplification.

\begin{algorithm}
	\caption{Training with a circular queue of environment instances.}
	\begin{algorithmic}[1]
		\Procedure{Train}{}
		
		\State EnvQueue $\gets [(N_1, D_1), \dots (N_{n-1}, D_{n-1})].$ // n GPU instances in total, 1 reserved for RL training
		\State policy $\pi$ $\gets$ random
		\State iters $\gets$ constant
		\State batch $\gets$ constant
		
		\For {iter in [ $0, \dots $iters]}		   		
		\For {b in [ $0, \dots $batch]}		  // distributed simulation
		\State CurrentEnv = EnvQueue.next()
		\State obs = CurrentEnv.reset()	   	    
		\For {$t$ in [$0, \dots l$]} // episode rollout
		\State action $\gets$ $\pi(\text{obs})$
		\State obs, reward = CurrentEnv.step(action)
		\EndFor	    
		\EndFor
		\State Update $\pi$
		\EndFor
		
		\State \textbf{return}  $\pi$ 
		\EndProcedure
	\end{algorithmic}
	\label{algo:case}
\end{algorithm}

\begin{algorithm}
	\caption{Step method of the environment}\label{algo:step}
	\begin{algorithmic}[1]
		\Procedure{Env.step}{}
		\State \textbf{arguments} actions
		\State Env.CurrentLayer.weights.prune(actions)
		\State accuracy, obs = Env.net.train()
		\State reward = Env.get\_reward(accuracy, actions)
		\State Env.CurrentLayer $\gets$ Env.next\_layer()
		\State \textbf{return} obs, reward
		\EndProcedure
	\end{algorithmic}
\end{algorithm}

The training our RL agents use a circular queue of environment instances as described in the body of the paper.
Assuming we have $n$ GPU intances in total, during each training the RL agent uses one GPU and the distributed simulaiton utilized the rest $n-1$,
with each episode taking one.
The environment is initialized with a circular queue of single $(N,D)$ instances.
After the completion of every episode the next network dataset pair $(N,D)$ in the queue is chosen and its last layer's observation $\Phi(L_t)$ is set as the initial state of the episode.
This way at each episode the agent trains against a new network and therefore learns to generalize. 
The learning algorithm of this environment is described in algorithm \ref{algo:case}.
A method of note is the step method of an MDP environment, which is described in algorithm \ref{algo:step}.

\subsection{Training performance}
We train multiple RL agents for different Compression Factor (CF); each policy is trained with 6 \texttt{p3.8xlarge} SageMaker instances. 
Figure \ref{fig:training_reward} plots the reward function of the RL policy in training as a function of training time (measured in hours). The shaded band around the mean line shows the minimum and maximum rewards during exploration.
\begin{figure}	
	\includegraphics[width=0.485\linewidth]{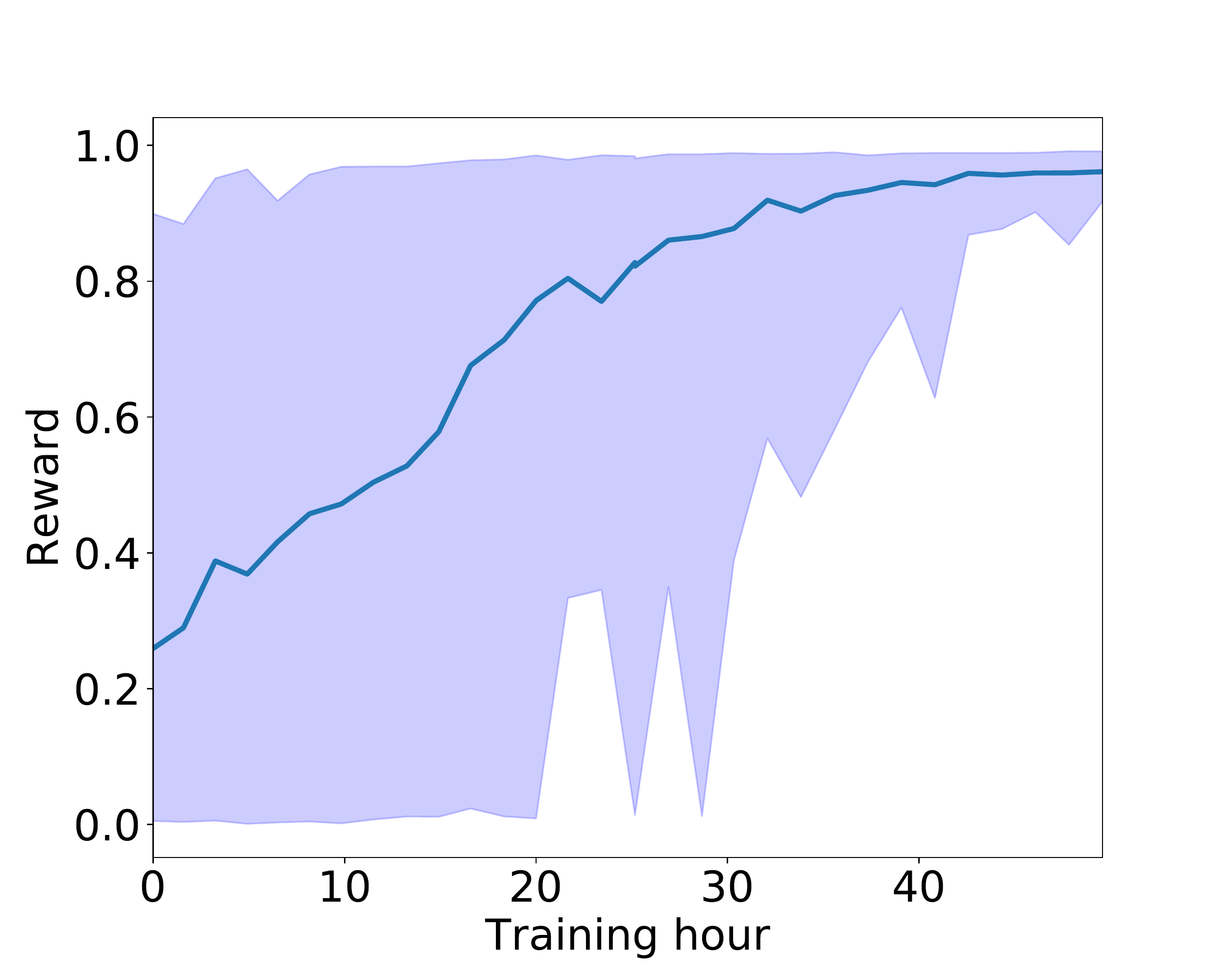}
	\includegraphics[width=0.485\linewidth]{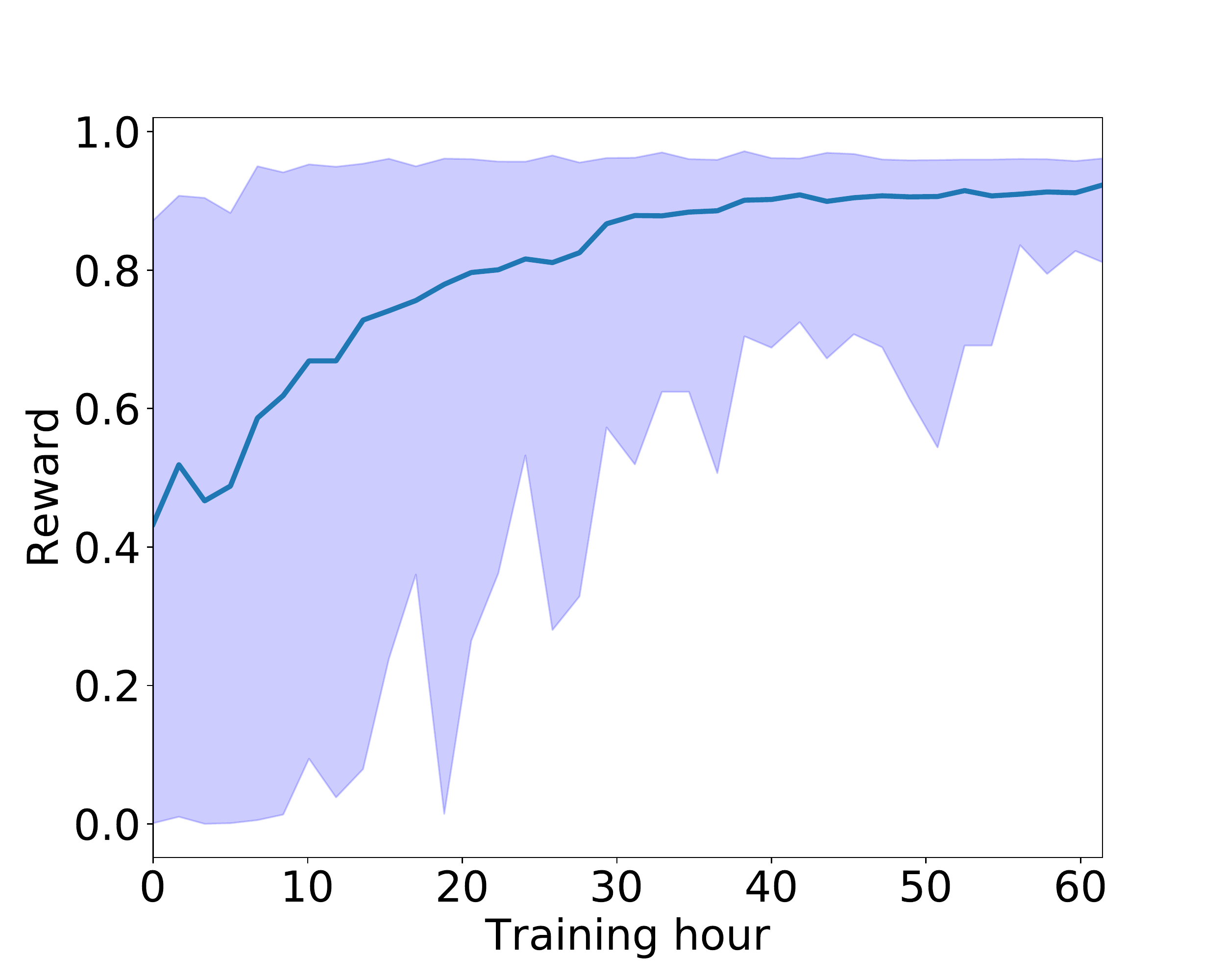}	 
	\caption{RL training curve for CF$=3$ (left panel) and CF$=6$ (right panel).}
	\label{fig:training_reward}	    
\end{figure}

\end{document}